\documentclass{article} 
\usepackage{iclr2027_conference,times}
\setcitestyle{numbers,square,comma}

\usepackage{amsmath,amsfonts,bm}

\def\eqref#1{equation~\ref{#1}}

\def\1{\bm{1}}

\DeclareMathAlphabet{\mathsfit}{\encodingdefault}{\sfdefault}{m}{sl}
\SetMathAlphabet{\mathsfit}{bold}{\encodingdefault}{\sfdefault}{bx}{n}

\usepackage{booktabs}
\usepackage{multirow}
\usepackage{array}
\usepackage{hyperref}
\usepackage{url}
\usepackage{booktabs}
\usepackage{makecell}
\usepackage[table]{xcolor}
\usepackage{graphicx}
\usepackage{subcaption}
\usepackage{multirow}
\usepackage{bbm}
\usepackage{xcolor}
\title{OnceSelect: Reusable Data Selection for Efficient Multimodal Instruction Tuning}
\iclrfinalcopy

\author{
Mingkang Dong$^{1}$\thanks{Equal contribution.},
Muxin Pu$^{2*}$,
Hongyi Cai$^{3}$,
Jie Li$^{5}$,
Jiancheng Pan$^{4}$,
Xu Zheng$^{6}$,
Yadan Luo$^{1}$,
Yuqian Fu$^{7}$\thanks{Corresponding author.}
\\[5pt]
$^{1}$The University of Queensland \qquad
$^{2}$Monash University \qquad
$^{3}$Universiti Malaya \quad
$^{4}$Tsinghua University \qquad
\\
$^{5}$Huazhong University of Science and Technology \quad
$^{6}$Great Bay University \qquad
$^{7}$KAUST
}

\makeatletter
\def\@noticestring{}
\makeatother
\begin{document}
\maketitle
\lhead{Preprint}
\begin{figure*}[h]
    \centering
    \vspace{-0.35in}
    \includegraphics[width=\textwidth]{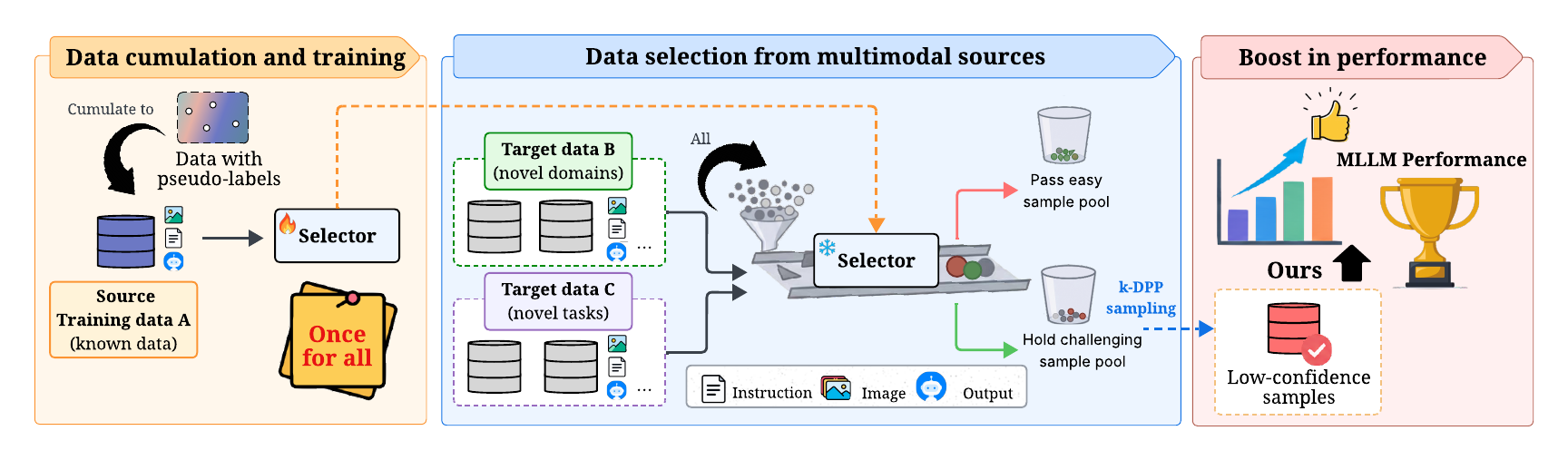}
    \vspace{-0.25in}
    \caption{\textbf{Motivation and overview of OnceSelect.}
A lightweight selector is trained once on cluster-derived pseudo-labels and
reused across multimodal instruction datasets without retraining. Within each
candidate dataset, low-confidence examples form candidate pools that are
further refined by cluster-wise $k$-DPP. The resulting subset can be reused
across downstream MLLMs without model-specific scoring or reselection.}
    \vspace{-0.15in}
    \label{fig:teaser}
\end{figure*}
\begin{abstract}

Multimodal instruction tuning is the de facto training method for adapting
multimodal large language models (MLLMs), yet the large-scale image-text
datasets it relies on are often highly redundant. Effective data selection is
therefore essential for efficient multimodal instruction tuning. However,
existing selection methods are often tightly coupled to the specific dataset,
target model, or training state used to derive their selection signals.
Consequently, applying them to a new dataset or downstream model typically
requires retraining the selector or recomputing model-dependent scores,
leading to substantial repeated computation. This raises a fundamental practical
question: \textit{can a selection signal be learned once and reused across datasets
and target models?} To this end, we propose \textbf{OnceSelect}, a reusable data selector for multimodal instruction tuning that is trained only once and then directly applied to unseen datasets. OnceSelect encodes image-instruction pairs in a frozen joint multimodal space, clusters them to obtain pseudo-labels that capture the coarse semantic structure, and trains a lightweight selector and retains an intermediate checkpoint selected according to a fixed validation macro-accuracy target. Samples receiving low confidence under this learned semantic partition are treated as potentially informative candidates. Because the selector is independent of the downstream MLLM, it can be reused
across candidate datasets without retraining or model-dependent scoring, while each selected subset can subsequently be reused across model architectures. Using only 15\% of LLaVA-625K, OnceSelect retains 99.2\% of full-data
aggregate performance across nine evaluation metrics. Without selector
retraining, it achieves 103.4\% and 102.5\% relative performance on the unseen
Vision-Flan-186K~\citep{VisionFlan} and LRV-Sub-180K~\citep{LRV},
respectively. Moreover, the same 93.7K LLaVA subset achieves
99.9--106.7\% relative performance across the evaluated Qwen3-VL and InternVL3
models without model-specific scoring or reselection. Together, these results establish OnceSelect as a reusable and efficient data selection framework across datasets and downstream model architectures. Code is available at https://github.com/DMK041218/OnceSelect

\end{abstract}

\section{Introduction}
\label{sec:introduction}

Multimodal large language models (MLLMs) acquire their instruction-following capabilities
by fine-tuning on large-scale multimodal datasets such as LLaVA-625K
\citep{llava}, Vision-Flan-186K \citep{VisionFlan}, and LRV-Sub-180K
\citep{LRV}. Despite their scale, these datasets often contain substantial
redundancy and weakly informative or incorrectly paired examples
\citep{VisNec}. Training on the complete data is therefore expensive and not
necessarily optimal: what matters is not only how much data is available, but
whether the selected samples provide informative and diverse supervision.

Existing data-selection methods estimate sample value using gradient influence \citep{icons}, training dynamics \citep{XMAS}, model losses \citep{VisNec, li2024quantityqualityboostingllm}, or learned quality scores
\citep{yan2025coidoefficientdataselection}. Although effective, their selection
signals are often tied to a particular target model, candidate dataset, or
training state. As a result, changing the dataset or downstream MLLMs may require
retraining the selector or recomputing model-dependent scores. This limitation
raises a practical question: 
\emph{can we learn a reusable selection criterion that remains effective across datasets and model architectures?}


Our premise is that different multimodal instruction datasets can share
\textit{recurring} image-instruction patterns. This suggests that the semantic structure learned from a source dataset can serve as a \textit{common} reference for data selection. Rather than estimating how much each sample will benefit a particular downstream model, we assess how readily it can be explained by this learned semantic structure. We capture this relation using the prediction confidence of a lightweight classifier trained on representative source examples. Our fundamental hypothesis is that low-confidence samples, which are less readily explained by the learned structure, may provide complementary supervision beyond readily classified examples.

Based on this insight, we propose \textbf{OnceSelect}, a reusable data selection framework for multimodal instruction tuning. As illustrated in Fig.~\ref{fig:teaser}, we first cluster frozen multimodal representations to construct cluster-derived pseudo-labels from a source training dataset, and then train a lightweight selector on centroid-nearest source examples, giving each cluster equal weight during training. 
Rather than using a fully converged classifier, we retain an intermediate
checkpoint selected according to a fixed validation macro-accuracy target. This allows the selector to capture broad semantic regularities while preserving informative variation in sample confidence. We then prioritize low-confidence candidates within each cluster, as these samples are less readily explained by the learned semantic structure and may provide complementary supervision.
%
To prevent uncertain candidates from concentrating in narrow or redundant
regions, OnceSelect combines selector uncertainty with cluster-aware \(k\)-DPP
sampling. Proportional cluster allocation maintains coarse semantic coverage,
while the \(k\)-DPP kernel assigns larger quality weights to lower-confidence candidates
and discourages the joint selection of semantically redundant examples. The
resulting subset therefore emphasizes difficult and informative samples while
remaining broadly representative of the candidate distribution.

Crucially, the learned semantic reference can be reused beyond its source
dataset. Samples from a new candidate dataset are mapped into the same
representation space and evaluated relative to the source-learned structure,
without target-specific pseudo-label construction, selector retraining, or
inference through the downstream MLLM. Once a subset has been selected, the
same data can also be used to fine-tune different MLLM architectures without
model-specific rescoring or reselection. Experiments show that OnceSelect preserves
99.2\% of full-data performance using only 15\% of LLaVA-625K, transfers
effectively to the unseen 
Vision-Flan-186K~\citep{VisionFlan} and LRV-Sub-180K~\citep{LRV} datasets.
and
generalizes across LLaVA~\citep{llava}, Qwen3-VL~\citep{qwen3}, and InternVL3~\citep{internvl}. 

Our contributions are summarized as follows:
\begin{itemize}
    \item We propose OnceSelect, a reusable data selection framework
    that learns a semantic reference from a large, heterogeneous
    multimodal dataset. The selector is trained once and reused
    across unseen datasets without selector retraining or
    target-model-specific rescoring.

    \item 
    We investigate low confidence from a partially converged selector as a proxy for sample informativeness, prioritizing samples that remain difficult to classify into the learned semantic clusters without requiring downstream-model gradients or benchmark supervision.

    \item We combine uncertainty-aware candidate selection with
    cluster-aware $k$-DPP sampling to balance informativeness,
    diversity, and semantic coverage. Experiments show competitive
    performance using a small fraction of the training data,
    supporting selector reuse across the evaluated datasets
    and selected-subset reuse across MLLM architectures.
\end{itemize}

\section{Related Work}
\subsection{Multimodal Instruction Tuning}
To adapt pre-trained MLLMs to downstream tasks~\citep{Pistachio,zhu2026egosound, ObjectStream, fu2025objectrelator, pan2026v, Evo-Depth,AutoDebias} and domain-specific scenarios~\citep{li2026egocross, balauca2025understanding, zhang2026egonight}, instruction tuning has become the standard practice, fine-tuning a pretrained backbone on large collections of image instruction-response pairs so that it learns to follow user intent \citep{llava}. As this paradigm has matured, however, the scale of instruction datasets has grown rapidly, with collections such as LLaVA-625K \citep{llava} and Vision-Flan \citep{VisionFlan} reaching hundreds of thousands of samples, and the training time and computational resources demanded by such corpora have become a pressing concern. Moreover, recent studies observe that these datasets contain a large portion of redundant and even incorrectly paired image-text samples \citep{VisNec}, causing the marginal benefit of additional data to diminish as the dataset grows. Selecting a subset of the most valuable and diverse samples has therefore become crucial for efficient and effective instruction tuning.
\subsection{Data Selection for Instruction Tuning}
To improve data efficiency, a growing body of work studies how to select a compact yet informative and diverse subset from large instruction datasets. Existing efforts can be broadly divided into gradient-based, scoring-based, and representation-based approaches.\\
\textbf{Gradient-based methods.} Gradient-based methods estimate the importance of training samples by analyzing their influence on the model's loss or parameter updates. Representative approaches, such as ICONS \citep{icons} and XMAS \citep{XMAS}, track gradient alignment or training dynamics to identify samples that maximize the model's learning trajectory. While these methods are theoretically well-grounded and provide high-fidelity importance scores, they suffer from a severe practical limitation: the influence signal is inherently coupled to the specific model architecture and its current parameter state. Consequently, whenever the target model, the task objective, or even the initial checkpoint changes, these signals must be rescored from scratch. Performing per-sample back-propagation over a massive candidate pool for every new training configuration is computationally prohibitive, 
limiting scalability in multimodal instruction tuning.

\noindent\textbf{Scoring-based methods.} Scoring-based methods assign a scalar quality or difficulty score to each sample and retain those that are top-ranked. These scores are typically derived in two ways. One approach aggregates multiple signals, such as model loss, perplexity, instruction complexity, or visual difficulty, into a unified quality estimate. For instance, Self-Filter \citep{chen-etal-2024-vision} leverages a learned scoring network to fuse feature-based criteria for instruction ranking. VisNec \citep{VisNec} measures the visual necessity to rank each sample. Another approach like CoIDO \citep{yan2025coidoefficientdataselection} queries external APIs to rate sample quality or difficulty.

\noindent\textbf{Representation-based and training-free methods.}
Recent work increasingly explores selection directly in representation space
to reduce the cost of model-dependent scoring. RDS and its scalable variant
RDS+ perform selection using pretrained hidden representations and show that
simple representation-based criteria can remain competitive even when scaling to very large instruction pools. PRISM further proposes a training-free multimodal selection strategy that exploits intrinsic properties of visual
representations, using correlation-based criteria to identify high-value
training samples without proxy-model training or gradient computation.
Similarly, ScalSelect~\citep{scalselect} avoids gradient computation and
constructs instruction-conditioned visual representations from the target MLLM to preserve the dominant structure of the full dataset.
DOSE~\citep{dose} leverages off-the-shelf models to assess data quality and image--text alignment, while DICS~\citep{dics} further considers intrinsic multimodal consistency and diversity for visual instruction selection.

Despite their improved efficiency, these methods still require computing
selection-specific representations or statistics for each new candidate
dataset, and some of them remain dependent on the target model used to extract the representations. 
In contrast, OnceSelect trains a lightweight selector once in a frozen multimodal
representation space and reuses it across different datasets and target
architectures. 
\begin{figure*}[t]
\centering
\includegraphics[width=\textwidth]{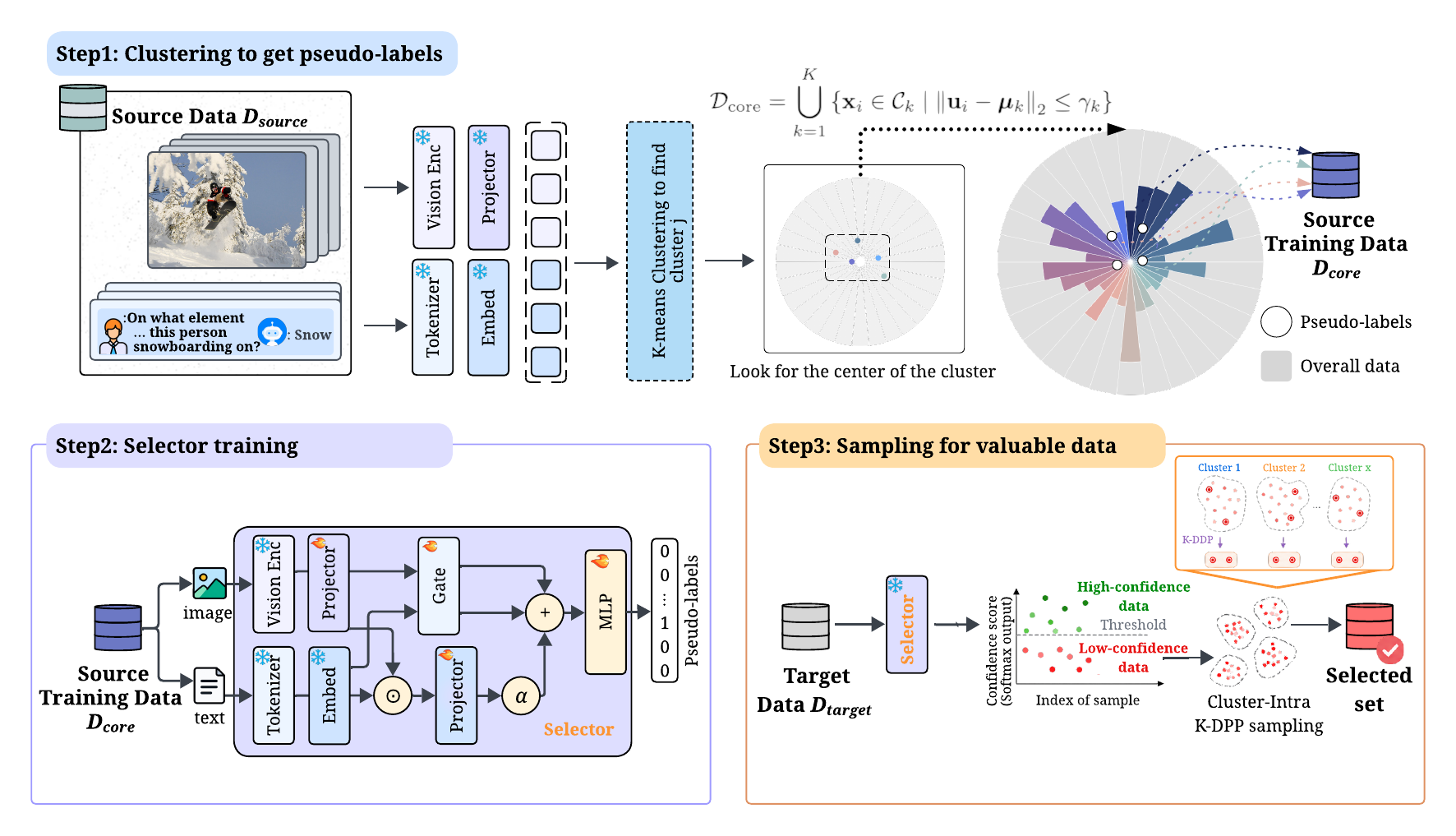}
\caption{\textbf{Overview of OnceSelect.}
(1) Frozen multimodal representations from a source dataset are clustered to construct a pseudo-labeled core from centroid-nearest samples.
(2) A lightweight selector is trained on this core with dynamic modality gating and cross-modal interaction.
(3) The frozen selector is applied to a candidate dataset to identify low-confidence samples, followed by cluster-wise $k$-DPP sampling to balance informativeness, diversity, and semantic coverage.}

\label{fig:methodology}
\end{figure*}

\section{Methodology}
\label{sec:method}

\subsection{Setup}
\label{sec:setup}
Given a source instruction-tuning dataset
$\mathcal{D}_{\mathrm{source}}
=\{\mathbf{x}_i\}_{i=1}^{N}$,
our framework selects a compact subset
$\mathcal{D}_{\mathrm{filtered}}
\subseteq\mathcal{D}_{\mathrm{source}}$.
Each sample $\mathbf{x}_i=(v_i,t_i,o_i)$ consists of an image $v_i$, an
instruction $t_i$, and a target response $o_i$. Only the image--instruction
pair $(v_i,t_i)$ is used during data selection, while the target response
$o_i$ is carried along unchanged for downstream fine-tuning.

Given a selection ratio $\rho$, we allocate the global budget across $K$
semantic clusters and construct
\begin{equation}
\mathcal{D}_{\mathrm{filtered}}
=
\bigcup_{k=1}^{K}\mathcal{S}_k,
\qquad
\mathcal{S}_k\subseteq\mathcal{C}_k,
\qquad
\sum_{k=1}^{K}|\mathcal{S}_k|
=
\left\lceil\rho N\right\rceil,
\label{eq:selection_overview}
\end{equation}
where $\mathcal{C}_k$ denotes the $k$-th semantic cluster and
$\mathcal{S}_k$ is the subset selected from that cluster. The quota assigned
to each cluster is proportional to its population, thereby maintaining the
coarse-grained semantic distribution of the original dataset.

Our selection criterion jointly considers informativeness and diversity.
Informativeness is estimated from the uncertainty of a lightweight selector,
while diversity is encouraged using an uncertainty-weighted $k$-DPP.
Unless otherwise specified, we retain $\rho=15\%$ of the candidate dataset.

\subsection{Multimodal Representation and Clustering}
\label{sec:cluster}

\noindent\textbf{Multimodal representation.}
For each sample, we use a frozen CLIP ViT-B/32~\citep{clipscore} encoder to extract image and
instruction representations. Let
$\mathbf{e}_i^v,\mathbf{e}_i^t\in\mathbb{R}^{d}$ denote the corresponding
$\ell_2$-normalized embeddings. For a multi-turn conversation, we encode each
valid user turn independently, average the normalized turn-level embeddings,
and normalize the resulting representation to obtain $\mathbf{e}_i^t$.
We then construct the joint multimodal representation as
\begin{equation}
\mathbf{u}_i
=
\operatorname{Norm}
\left(
[\mathbf{e}_i^v;\mathbf{e}_i^t]
\right),
\label{eq:joint_representation}
\end{equation}
where $[\cdot;\cdot]$ denotes concatenation and
$\operatorname{Norm}(\mathbf{x})=\mathbf{x}/\|\mathbf{x}\|_2$.
This joint representation captures both visual content and instruction
semantics while remaining independent of the downstream models.

\noindent\textbf{Semantic clustering.}
We partition the source dataset into $K$ semantic clusters by applying
MiniBatch K-Means~\citep{kmeans} to the joint representations:
\begin{equation}
\min_{\{a_i\}_{i=1}^{N},\,\{\boldsymbol{\mu}_k\}_{k=1}^{K}}
\sum_{i=1}^{N}
\left\|
\mathbf{u}_i-\boldsymbol{\mu}_{a_i}
\right\|_2^2,
\qquad
a_i\in\{1,\ldots,K\},
\label{eq:clustering}
\end{equation}
where $a_i$ denotes the cluster assignment of sample $i$, and
$\boldsymbol{\mu}_k$ denotes the centroid of cluster $k$. The resulting
clusters provide a coarse partition of the source distribution and serve
as pseudo-labels for training the transferable selector.

\subsection{Clustering-Based Pseudo-Labeling}
\label{sec:pseudo}

Samples close to a cluster centroid provide representative examples of the
corresponding semantic region. We therefore construct a pseudo-labeled coreset using the centroid-nearest samples:
\begin{equation}
\mathcal{D}_{\mathrm{core}}
=
\bigcup_{k=1}^{K}
\left\{
\mathbf{x}_i\in\mathcal{C}_k
\;\middle|\;
\left\|
\mathbf{u}_i-\boldsymbol{\mu}_k
\right\|_2
\leq\gamma_k
\right\},
\label{eq:core}
\end{equation}
Each retained sample inherits the index $k$ of its source cluster as its
pseudo-label. Here, $\gamma_k$ is a cluster-specific distance threshold chosen
to retain the centroid-nearest fraction $\eta$ of $\mathcal{C}_k$. Ties at
the selection boundary are resolved using a deterministic ordering. We set $\eta=3\%$ by default. Within each cluster, the retained core samples
are split into 80\% for selector training and 20\% for validation. The unions
of the resulting cluster-wise splits are denoted by
$\mathcal{D}_{\mathrm{core}}^{\mathrm{tr}}$ and
$\mathcal{D}_{\mathrm{core}}^{\mathrm{val}}$, respectively.

\subsection{Selector Training}
\label{sec:train}

The selector is trained to predict the cluster-derived pseudo-labels from
separate image and instruction representations. A conventional
concatenation-based fusion module applies a fixed fusion mechanism across
samples and does not explicitly adapt the relative contribution of each
modality to the visual dependence of individual instructions. We therefore
introduce a dynamic gated fusion module.
Specifically, the image and instruction features are first projected through
modality-specific branches. A sample-dependent scalar gate is then predicted
from the two modalities and their elementwise interaction:
\begin{equation}
\begin{aligned}
\mathbf{z}_i
&=
\mathbf{e}_i^v\odot\mathbf{e}_i^t,\\
g_i
&=
\sigma\!\left(
g_{\psi}
\left(
[\mathbf{e}_i^v;\mathbf{e}_i^t;\mathbf{z}_i]
\right)
\right),\\
\mathbf{h}_i
&=
g_i\phi_v(\mathbf{e}_i^v)
+
(1-g_i)\phi_t(\mathbf{e}_i^t)
+
\alpha\phi_{\mathrm{int}}(\mathbf{z}_i),
\end{aligned}
\label{eq:gated_fusion}
\end{equation}
where $g_i\in(0,1)$ is a sample-dependent scalar gate,
$\phi_v,\phi_t,\phi_{\mathrm{int}}:
\mathbb{R}^{d}\rightarrow\mathbb{R}^{h}$ are learnable projection branches,
and $\alpha$ controls the contribution of the cross-modal interaction
residual.
The first two terms dynamically balance visual and instructional information
for each sample. The interaction branch explicitly models dimension-wise
image--instruction interactions, allowing their correspondence to influence
the fused representation even when one modality receives a smaller gate
value.

A lightweight classification head maps $\mathbf{h}_i$ to a distribution over
the cluster-derived pseudo-labels:
\begin{equation}
\mathbf{p}_i
=
\operatorname{softmax}
\left(
f_{\theta}(\mathbf{h}_i)
\right).
\label{eq:selector_prediction}
\end{equation}
The selector is optimized on the training split of the pseudo-labeled core
using cross-entropy:
\begin{equation}
\mathcal{L}_{\mathrm{sel}}
=
-\frac{1}
{|\mathcal{D}_{\mathrm{core}}^{\mathrm{tr}}|}
\sum_{\mathbf{x}_i\in\mathcal{D}_{\mathrm{core}}^{\mathrm{tr}}}
\log p_{i,a_i}.
\label{eq:selector_loss}
\end{equation}
During optimization, mini-batches are constructed using inverse-class-frequency sampling to mitigate cluster imbalance. The selector is not intended to become a fully converged semantic classifier.
Instead, we retain a checkpoint whose validation macro accuracy is closest to
a predefined target. At this stage, the selector has learned the dominant
structure of the cluster-derived task without completely eliminating
uncertainty on ambiguous or atypical samples. We use this residual uncertainty
as a proxy for sample-level informativeness. Checkpoint details are in Appendix.~\ref{app:checkpoint}

Importantly, checkpoint selection depends exclusively on the pseudo-label
validation set. It does not use downstream fine-tuning performance or
benchmark annotations.

\subsection{Data Selection}
\label{sec:select}

The trained selector is then directly applied to calculate the confidence for candidate samples:
\begin{equation}
F(\mathbf{x}_i)
=
\max_{c\in\{1,\ldots,K\}}p_{i,c}.
\label{eq:confidence}
\end{equation}
A high value indicates that the selector can confidently associate a sample
with one of the learned semantic clusters. Conversely, a low value indicates
ambiguity under the learned cluster decision function and is treated as a
proxy for potential sample-level informativeness.

Cluster membership is determined independently of the selector confidence.
Specifically, each candidate is assigned to its nearest semantic centroid:
\begin{equation}
\hat{a}_i
=
\arg\min_{k\in\{1,\ldots,K\}}
\left\|
\mathbf{u}_i-\boldsymbol{\mu}_k
\right\|_2^2.
\label{eq:cluster_assignment}
\end{equation}
The assigned cluster $\hat{a}_i$ is used for quota allocation and
cluster-wise diversity selection, whereas the selector output is used only to
compute the confidence in ~\eqref{eq:confidence}. For samples in the source
dataset, this assignment coincides with the clustering result in
~\eqref{eq:clustering}. For an unseen candidate dataset, samples are
directly assigned to the semantic centroids learned from the source dataset.

\noindent\textbf{Low-confidence candidate construction.}
The global selection budget is distributed across the assigned clusters in
proportion to their populations. Let $q_k$ denote the final quota allocated to
cluster $\mathcal{C}_k$. Within each cluster, samples are ranked in ascending
order of selector confidence. We construct a candidate pool
$\mathcal{P}_k$ by retaining up to $2q_k$ samples with the lowest confidence,
providing a broader uncertain region from which the final $q_k$ samples are
selected.

Low-confidence ranking alone may retain redundant examples. For example,
samples located near the same decision boundary may have similarly low
confidence while conveying nearly identical multimodal information. We
therefore combine uncertainty-based informativeness with determinantal
diversity.

\noindent\textbf{Uncertainty-weighted diversity selection.}
For each candidate pool $\mathcal{P}_k$, let
$r_i\in\{1,\ldots,|\mathcal{P}_k|\}$ denote the rank of sample $i$ when
candidates are ordered by ascending confidence. We define its rank-based
informativeness weight as
\begin{equation}
w_i
=
1-
\frac{r_i-1}{|\mathcal{P}_k|},
\qquad
i\in\mathcal{P}_k,
\label{eq:informativeness_weight}
\end{equation}
such that samples with lower selector confidence receive larger weights.
The rank-based formulation avoids relying on the absolute calibration of the
selector probabilities.
Using the normalized joint representations from
~\eqref{eq:joint_representation}, we construct a separate $L$-ensemble
kernel for each cluster-specific candidate pool:
\begin{equation}
L_{ij}^{(k)}
=
\sqrt{w_iw_j}
\left(
\frac{1+\mathbf{u}_i^\top\mathbf{u}_j}{2}
\right),
\qquad
i,j\in\mathcal{P}_k.
\label{eq:dpp_kernel}
\end{equation}
The informativeness weights favor uncertain samples, while the kernel
determinant discourages redundant selections~\citep{kdpp}.
For scalability, we approximate cluster-level $k$-DPP sampling by
partitioning each candidate pool into blocks of at most 128 samples
after a fixed-seed shuffle. We allocate the cluster quota $q_k$
proportionally across blocks and perform exact fixed-cardinality
$k$-DPP sampling within each block, preserving the cluster quotas
and global selection budget.

Together, proportional cluster allocation preserves coarse-grained
semantic coverage, selector uncertainty prioritizes potentially
informative samples, and $k$-DPP reduces local redundancy.
These complementary stages balance representativeness,
informativeness, and diversity under a fixed data budget.
\section{Experiments and Results}
\label{sec:exp}
\subsection{Experimental Setup}
\label{sec:setup}
\noindent\textbf{Datasets.}
We evaluate OnceSelect on three multimodal instruction-tuning datasets:
LLaVA-625K~\citep{llava}, Vision-Flan-186K~\citep{VisionFlan}, and
LRV-Sub-180K~\citep{LRV}. We use LLaVA-625K as the source dataset for
training the selector and evaluating in-domain selection. The trained selector
is then directly applied to Vision-Flan-186K and LRV-Sub-180K as unseen target
datasets, without retraining or target-specific pseudo-label construction. Details are in Appendix.~\ref{app:datasets}

\noindent\textbf{Baselines.}
We compare against Random, Length, Perplexity \citep{perplexity},
COINCIDE \citep{lee2024coincide}, PRISM \citep{prism},
RDS+ \citep{rds}, ScalSelect \citep{scalselect}, XMAS \citep{XMAS} and
VisNec \citep{VisNec}. These baselines cover heuristic,
representation-based, and multimodal-specific selection strategies. Baseline details are in Appendix.~\ref{app:baselines}

\noindent\textbf{Evaluation Benchmarks.}
We report nine metrics across benchmark suites:
MMBench-En/Cn \citep{liu2023mmbench}, MME-P/C \citep{MME},
AI2D \citep{ai2d}, SQA-IMG \citep{sqa}, POPE-A/P \citep{pope},
and OCRBench \citep{ocrbench}. For MMBench, we report circular accuracy on the English and Chinese splits. For POPE, we report accuracy (POPE-A) and precision
(POPE-P), each macro-averaged over the random, popular, and adversarial subsets. Details are in Appendix.~\ref{app:evaluation_benchmarks}

\noindent\textbf{Selector Training Details.}
We extract $\ell_2$-normalized multimodal features using frozen CLIP
ViT-B/32~\citep{clipscore} and apply MiniBatch K-means~\citep{kmeans} with $K=128$. The gated selector is
trained on the centroid-nearest pseudo-labeled core using AdamW with a learning
rate of $1\times10^{-5}$. Details are in Appendix.~\ref{app:selector_details},~\ref{app:checkpoint},,~\ref{app:kdpp}

\noindent\textbf{MLLMs Instruction Tuning Details.}
For a fair and reproducible comparison, we follow the standard LLaVA fine-tuning pipeline throughout,
keeping the model architecture, hyperparameters, and optimization schedule. We perform LoRA~\citep{hu2022lora} fine-tuning for one epoch. Details are in Appendix.~\ref{app:llava_training}.
\subsection{Main Results on Multimodal Data Selection}
As shown in Tab.~\ref{tab:main_result}, we compare different multimodal data
selection methods on LLaVA-625K under the same 15\% data budget. In this
in-domain setting, LLaVA-625K serves as both the source dataset for learning
the selector and the candidate pool for data selection. Specifically, we form
the pseudo-labeled core set using the centroid-nearest 3\% of samples within
each semantic cluster and train the selector only on this core set. The
resulting selector is then applied to the complete LLaVA-625K candidate pool
to select 15\% of the samples for instruction-tuning LLaVA-v1.5-7B.
Fine-tuning on the complete 625K dataset serves as the full-data reference.

Our method achieves the highest overall relative performance of 99.2\%,
outperforming the recent baseline, ScalSelect, by 1.6 percentage points and
XMAS by 2.9 points. It obtains the best results among all selection methods on
MME-C, AI2D, POPE-A, and OCRBench. Notably, it surpasses full-data training on
MME-C and POPE-A, while nearly matching
it on AI2D (53.7 vs.\ 53.8). Other methods lead individual metrics, including COINCIDE on the two MMBench
splits and Length on POPE-P, whereas OnceSelect provides the strongest
aggregate result across nine reported metrics.
\begin{table*}[t]
\centering
\caption{Comparison with existing multimodal data selection methods for
LLaVA-v1.5-7B on LLaVA-625K. \textbf{Bold} indicates the best result among
data selection methods.}
\LARGE
\label{tab:main_result}
\renewcommand{\arraystretch}{1.15}
\resizebox{\textwidth}{!}{%
\begin{tabular}{l|ccccccccc|c}
\toprule
\textbf{Method}
& \textbf{MMBench-En}
& \textbf{MMBench-Cn}
& \textbf{MME-P}
& \textbf{MME-C}
& \textbf{AI2D}
& \textbf{POPE-A}
& \textbf{POPE-P}
& \textbf{SQA-IMG}
& \textbf{OCRBench}
& \textbf{Rel.} \\
\midrule

Full-Finetune
& 64.0 & 56.7 & 1463.9 & 278.6 & 53.8
& 85.2 & 93.1 & 67.6 & 31.3 & 100.0\% \\
\midrule

Random
& 58.0 & 51.7 & 1418.9 & 295.4 & 50.8
& 85.0 & 90.7 & 65.2 & 27.8 & 95.7\% \\

Length
& 49.1 & 36.4 & 1283.0 & 292.1 & 37.4
& 81.8 & \textbf{96.7} & 52.7 & 24.7 & 84.4\% \\

Perplexity
& 31.7 & 32.5 & 1403.1 & 278.9 & 38.0
& 84.8 & 93.0 & 48.4 & 26.1 & 80.9\% \\

COINCIDE {\small (EMNLP 2024)}
& \textbf{62.2} & \textbf{55.4} & \textbf{1422.0}
& 274.3 & 51.7 & 83.8 & 91.0
& \textbf{66.8} & 28.7 & 97.0\% \\

PRISM {\small (EMNLP 2026)}
& 61.5 & 53.0 & 1396.2 & 276.4 & 50.9
& 83.4 & 93.3 & 64.1 & 28.6 & 95.9\% \\

RDS+ {\small (arXiv 2025)}
& 56.1 & 50.2 & 1414.7 & 302.9 & 50.8
& 84.8 & 92.2 & 64.5 & 29.4 & 96.0\% \\

ScalSelect {\small (ACMMM 2026)}
& 59.2 & 52.8 & 1400.3 & 308.6 & 51.8
& 84.0 & 94.7 & 65.3 & 29.3 & 97.6\% \\

VisNec {\small (ECCV 2026)}
& 58.6 & 52.7 & 1390.3 & 298.2 & 49.8
& 86.5 & 92.4 & 62.5 & 30.1 & 97.0\% \\

XMAS {\small (ICML 2026)}
& 61.9 & 54.3 & 1356.2 & 300.7 & 50.0
& 84.3 & 90.9 & 60.3 & 29.6 & 96.3\% \\

\rowcolor{blue!10}
Ours
& 60.0 & 53.4 & 1420.6
& \textbf{314.6} & \textbf{53.7} & \textbf{86.9}
& 92.9 & 63.9 & \textbf{30.9} & \textbf{99.2\%} \\

\bottomrule
\end{tabular}%
}
\vspace{-0.15in}
\end{table*}
\subsection{Transferability to Unseen Datasets}
\label{sec:transfer-data}
Once the selector is trained on LLaVA-625K, we directly apply it to Vision-Flan-186K and LRV-Sub-180K without selector retraining.
Target samples are assigned
to the source semantic clusters using frozen CLIP representations.
We allocate quotas according to the target cluster distribution and
select 15\% of each dataset through low-confidence candidate construction
and blockwise $k$-DPP, without downstream evaluation supervision. As shown in Tab.~\ref{tab:unseendataset}, the frozen source selector reaches
103.4\% relative performance on Vision-Flan using 27.9K examples and 102.5\%
on LRV-Sub using 27.1K examples. On LRV-Sub, OnceSelect improves over
ScalSelect by 1.1 and 8.5 points on MMBench-En and MMBench-Cn, respectively,
and exceeds the full-data reference on both splits. These results support
selector reuse across the two evaluated target distributions.
\begin{table*}[h]
\centering
\vspace{-10pt}
\caption{Overall performance and efficiency comparison of selection approaches for
fine-tuning LLaVA-v1.5-7b on Vision-Flan-186K and LRV-Sub-180K datasets using a 15\%  sampling ratio across various multimodal evaluation benchmarks. The best result is \textbf{bolded}.}
\huge
\label{tab:unseendataset}
\renewcommand{\arraystretch}{1.15}
\resizebox{\textwidth}{!}{%
\begin{tabular}{
l|l|c|
c c c c c c c c c|c
}
\toprule
\textbf{Dataset}
& \textbf{Method}
& \textbf{\#Samples}
& \textbf{MMBench-En}
& \textbf{MMBench-Cn}
& \textbf{MME-P}
& \textbf{MME-C}
& \textbf{AI2D}
& \textbf{POPE-A}
& \textbf{POPE-P}
& \textbf{SQA-IMG}
& \textbf{OCRBench}
& \textbf{Rel.} \\
\midrule

\multirow{4}{*}{LRV-Sub-180K}
& Full-Finetune
& 180K
& 45.6
& 31.2
& 1082.2
& 258.9
& 41.0
& 75.4
& 69.1
& 58.6
& 27.8
& 100.0\% \\

& Random
& 27K
& 45.1
& 30.6
& 837.7
& 223.9
& 45.3
& 60.2
&\textbf{ 63.4}
& 56.1
& 27.5
& 93.1\% \\

& ScalSelect
& 27K
& 47.2
& 30.5
& 840.6
& 281.8
& 43.4
& 65.0
& 62.7
& \textbf{60.8}
& 27.4
& 97.0\% \\

\rowcolor{blue!10}
& \textbf{Ours}
& 27K
& \textbf{48.3}
& \textbf{39.0}
& \textbf{923.9}
& \textbf{289.6}
& \textbf{45.5}
& \textbf{67.6}
& 61.1
& 60.1
& \textbf{28.5}
& \textbf{102.5\%} \\
\midrule

\multirow{3}{*}{Vision-Flan-186K}
& Full-Finetune
& 186K
& 53.4
& 48.2
& 1238.1
& 245.3
& 37.7
& 84.9
& 89.2
& 55.6
& 26.7
& 100.0\% \\

& Random
& 27K
& 43.6
& 43.1
& \textbf{1238.6}
& \textbf{259.7}
& 35.8
& \textbf{85.0}
& \textbf{90.7}
& 62.1
& 27.8
& 98.8\% \\
\rowcolor{blue!10}
& Ours
& 27K
& \textbf{52.5}
& \textbf{48.3}
& 1179.7
& 256.8
& \textbf{38.3}
& 82.8
& 90.5
& \textbf{67.2}
& \textbf{29.5}
& \textbf{103.4\%} \\
\bottomrule
\end{tabular}}
\vspace{-0.15in}
\end{table*}
\subsection{Efficiency Analysis}
\begin{figure*}[h]
    \centering
    \includegraphics[width=\textwidth]{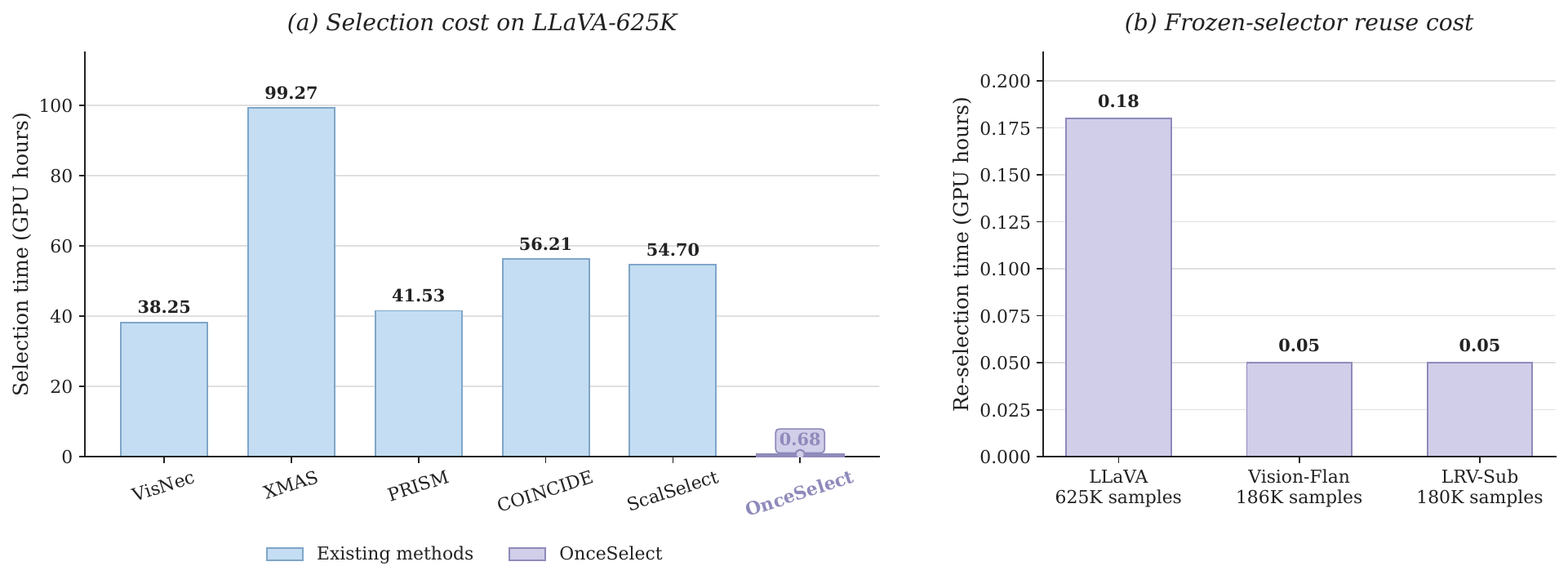}
    \caption{\textbf{Computational costs of data selection in GPU hours.}
(a) Comparison on LLaVA-625K, including one-time selector
training for OnceSelect.
(b) OnceSelect selection cost with the frozen selector.
All costs are measured on a single NVIDIA A6000 GPU.}
    \label{fig:efficiency}
\vspace{-0.2in}
\end{figure*}

Fig.~\ref{fig:efficiency} compares data selection costs.
In Panel (a), OnceSelect requires 0.68 GPU hours on LLaVA-625K,
including one-time selector training and selection.
After training, the frozen selector can be reused without retraining.
As shown in Panel (b), subsequent selection costs are approximately
0.18 GPU hours for LLaVA-625K and 0.05 GPU hours each for
Vision-Flan-186K and LRV-Sub-180K. These results highlight the
efficiency of selector reuse.
\subsection{Transferability across Model Architectures}
\label{sec:transfer-architecture}
We fine-tune Qwen3-VL-4B, Qwen3-VL-8B, and InternVL3-8B using
the same 93.7K samples selected from LLaVA-625K, without selector
retraining or model-specific reselection. As shown in Tab.~\ref{tab:scalselect_comparison}, our subset achieves
106.7\% and 105.7\% relative performance on the two Qwen3-VL models,
outperforming full-data training, Random, and ScalSelect in aggregate.
MMBench-En/Cn reaches 85.7/82.7 and 86.8/85.4, respectively,
while OCRBench improves to 84.2 and 85.8 from the full-data
results of 69.1 and 66.0. On InternVL3-8B, it achieves 99.9\%
relative performance, nearly matching full-data training with only
15\% of the samples. These results support reusing the selected
subset across the evaluated model families and scales.

\begin{table*}[h]
\centering
\caption{Cross-architecture transfer of data selected from LLaVA-625K.
The same 93.7K subset is used across all target models without model-specific
scoring or reselection.}
\LARGE
\label{tab:scalselect_comparison}
\renewcommand{\arraystretch}{1.15}
\resizebox{\textwidth}{!}{%
\begin{tabular}{
l|l|c|
c c c c c c c c c|c
}
\toprule
\textbf{Model}
& \textbf{Method}
& \textbf{\#Samples}
& \textbf{MMBench-En}
& \textbf{MMBench-Cn}
& \textbf{MME-P}
& \textbf{MME-C}
& \textbf{AI2D}
& \textbf{POPE-A}
& \textbf{POPE-P}
& \textbf{SQA-IMG}
& \textbf{OCRBench}
& \textbf{Rel.} \\
\midrule

\multirow{4}{*}{Qwen3-VL-4B-Instruct}
& Full-Finetune
& 625K
& 82.1
& 71.5
& 1589.1
& 651.4
& 79.0
& 88.4
& 97.0
& 87.3
& 69.1
& 100.0\% \\

& ScalSelect
& 100K
& 84.3
& 81.4
& 1652.7
& 658.9
& 79.9
& 88.2
& 96.9
& 91.0
& 75.4
& 104.0\% \\

& Random
& 93.7K
& 81.3
& 70.3
& 1494.2
& 612.3
& 77.1
& 81.3
& 95.8
& 88.5
& 70.4
& 97.4\% \\

\rowcolor{blue!10}
& Ours
& 93.7K
& \textbf{85.7}
& \textbf{82.7}
& \textbf{1702.7}
& \textbf{659.6}
& \textbf{83.2}
& \textbf{89.0}
& 96.7
& \textbf{91.1}
& \textbf{84.2}
& \textbf{106.7\%} \\
\midrule

\multirow{4}{*}{Qwen3-VL-8B-Instruct}
& Full-Finetune
& 625K
& 84.3
& 82.7
& 1700.2
& 676.4
& 80.5
& 88.5
& 96.4
& 88.4
& 66.0
& 100.0\% \\

& ScalSelect
& 100K
& 82.5
& 82.1
& 1709.1
& 701.4
& 81.8
& 88.0
& 94.4
& \textbf{92.1}
& 77.3
& 102.4\% \\

& Random
& 93.7K
& 83.3
& 81.1
& 1695.2
& 655.9
& 76.9
& 85.5
& 90.1
& 86.3
& 69.7
& 98.0\% \\

\rowcolor{blue!10}
& Ours
& 93.7K
& \textbf{86.8}
& \textbf{85.4}
& \textbf{1738.8}
& \textbf{709.9}
& \textbf{84.8}
& \textbf{88.9}
& \textbf{95.2}
& 91.7
& \textbf{85.8}
& \textbf{105.7\%} \\
\bottomrule
\multirow{3}{*}{InternVL3-8B-Instruct}
& Full-Finetune
& 625K
& 85.7
& 86.1
& 1726.1
& 666.4
& 82.8
& 90.1
& 94.9
& 88.9
& 76.8
& 100.0\%\\

& Random
& 93.7K
& 84.2
& 83.5
& 1659.3
& 655.1
& 80.6
& 90.5
& 88.0
& 86.7
& 74.2
& 97.1\%\\

\rowcolor{blue!10}
& Ours
& 93.7K
& \textbf{85.4}
& \textbf{85.9}
& \textbf{1721.6}
& \textbf{661.1}
& \textbf{83.6}
& \textbf{91.0}
& \textbf{93.5}
& \textbf{89.4}
& \textbf{77.0}
& \textbf{99.9\%} \\
\bottomrule
\end{tabular}}
\end{table*}
\subsection{Ablation Study}
\label{sec:ablation}
As shown in Tab.~\ref{tab:component_ablation}, we investigate each component under the same
15\% data budget, with full-data training as a reference. The full method
achieves 99.2\% relative performance, compared with 97.0--97.4\% for
the ablated variants. Removing the dynamic gate degrades MME-P, MME-C,
and OCRBench despite improvements on MMBench-Cn and SQA-IMG,
suggesting benefits for overall balance rather than every metric.
Removing the interaction branch causes the largest drop on MME-C
(314.6 to 245.4), highlighting its contribution to cognition-oriented
tasks. Removing $k$-DPP lowers scores on both MME subsets, POPE,
and OCRBench, supporting diversity-aware selection beyond confidence
ranking.  Replacing the low-confidence pool with a high-confidence pool while retaining all other components reduces relative performance to 97.4\%, supporting uncertain samples as potentially informative candidates.
Although some ablations improve individual metrics, none matches
the full method overall, suggesting complementary benefits from
adaptive fusion, cross-modal interaction and k-DPP.
\begin{table*}[h]
\centering
\caption{Ablation study of the main components in our data selection framework.
Our full method and all ablation variants select 15\% (93.7K) of LLaVA-625K,
the best result is \textbf{bolded}.}
\LARGE
\label{tab:component_ablation}
\renewcommand{\arraystretch}{1.15}
\resizebox{\textwidth}{!}{%
\begin{tabular}{
l|c|
c c c c c c c c c|c
}
\toprule
\textbf{Variant}
& \textbf{\#Samples}
& \textbf{MMBench-En}
& \textbf{MMBench-Cn}
& \textbf{MME-P}
& \textbf{MME-C}
& \textbf{AI2D}
& \textbf{POPE-A}
& \textbf{POPE-P}
& \textbf{SQA-IMG}
& \textbf{OCRBench}
& \textbf{Rel.} \\
\midrule

Full Data Training
& 625K
& 64.0
& 56.7
& 1463.9
& 278.6
& 53.8
& 85.2
& 93.1
& 67.6
& 31.3
& 100.0\% \\
\midrule

w/o Dynamic Gate
& 93.7K
& 59.4
& 54.0
& 1362.0
& 293.2
& 53.2
& 85.9
& 92.5
& 65.6
& 28.7
& 97.1\% \\

w/o Interaction
& 93.7K
& \textbf{61.8}
& \textbf{54.7}
& 1404.7
& 245.4
& 53.3
& \textbf{87.2}
& 92.2
& 66.3
& 30.4
& 97.0\% \\

w/o k-DPP
& 93.7K
& 61.3
& 53.0
& 1361.2
& 271.8
& \textbf{54.4}
& 86.4
& 91.8
& \textbf{66.4}
& 29.6
& 97.1\% \\

High-Confidence Selection
& 93.7K
& 58.9
& 53.6
& 1418.3
& 291.4
& 51.6
& 85.4
& 94.1
& 65.4
& 29.7
& 97.4\%\\

\rowcolor{blue!10}
Full Method (Ours)
& 93.7K
& 60.0
& 53.4
& \textbf{1420.6}
& \textbf{314.6}
& 53.7
& 86.9
& \textbf{92.9}
& 63.9
& \textbf{30.9}
& \textbf{99.2\%}  \\

\bottomrule
\end{tabular}}
\vspace{-0.15in}
\end{table*}

We also conduct a sensitivity study on the selector backbone, the number of clusters, and selection ratio in Appendix.~\ref{app:transfer_protocols} 
\section{Conclusions}
In this work, we addressed a practical limitation of multimodal data
selection: methods that rely on model or dataset-specific selection
signals can require repeated scoring or adaptation when the training
setting changes. We proposed OnceSelect, a reusable data selection
framework that trains a lightweight selector once on a large source
dataset using frozen multimodal representations and cluster-derived
pseudo-labels. By combining uncertainty from a partially converged
selector with diversity-aware sampling, OnceSelect prioritizes potentially
informative examples while maintaining broad semantic coverage.
The frozen selector can be reused across candidate datasets without selector
retraining or downstream-model inference, while each selected subset can be
reused across model architectures without model-specific reselection. Across
the evaluated settings, OnceSelect retains competitive or better aggregate
performance using only 15\% of the original dataset.
\section{Limitations}
Our experiments use LLaVA-625K as the sole source dataset for constructing the
semantic clusters and pseudo-labeled selector-training core. We have not
systematically evaluated how the scale or diversity of the source corpus
affects selector quality and cross-dataset transfer. Future work will study
this scaling behavior and construct semantic references from multiple source
datasets.



\newpage 

\bibliographystyle{iclr2027_conference}
\bibliography{bib}

@misc{VisNec,
      title={VisNec: Measuring and Leveraging Visual Necessity for Multimodal Instruction Tuning}, 
      author={Mingkang Dong and Hongyi Cai and Jie Li and Sifan Zhou and Bin Ren and Kunyu Peng and Yuqian Fu},
      year={2026},
      eprint={2603.01195},
      archivePrefix={arXiv},
      primaryClass={cs.CV},
      url={https://arxiv.org/abs/2603.01195}, 
}

@inproceedings{internvl,
  title     = {InternVL: Scaling up Vision Foundation Models and Aligning
               for Generic Visual-Linguistic Tasks},
  author    = {Chen, Zhe and Wu, Jiannan and Wang, Wenhai and Su, Weijie and
               Chen, Guo and Xing, Sen and Zhong, Muyan and Zhang, Qinglong and
               Zhu, Xizhou and Lu, Lewei and Li, Bin and Luo, Ping and
               Lu, Tong and Qiao, Yu and Dai, Jifeng},
  booktitle = {Proceedings of the IEEE/CVF Conference on Computer Vision and
               Pattern Recognition (CVPR)},
  year      = {2024}
}

@inproceedings{zhu2026egosound,
  title={Egosound: Benchmarking sound understanding in egocentric videos},
  author={Zhu, Bingwen and Fu, Yuqian and Dong, Qiaole and Sun, Guolei and Qian, Tianwen and Wu, Yuzheng and Paudel, Danda Pani and Fu, Yanwei and Xue, Xiangyang},
  booktitle={Proceedings of the IEEE/CVF Conference on Computer Vision and Pattern Recognition},
  pages={25589--25598},
  year={2026}
}

@inproceedings{zhang2026egonight,
  title={Egonight: Towards egocentric vision understanding at night with a challenging benchmark},
  author={Zhang, Deheng and Fu, Yuqian and Yang, Runyi and Miao, Yang and Qian, Tianwen and Zheng, Xu and Sun, Guolei and Chhatkuli, Ajad and Huang, Xuanjing and Jiang, Yu-Gang and others},
  booktitle={International Conference on Learning Representations},
  volume={2026},
  pages={887--901},
  year={2026}
}

@inproceedings{fu2025objectrelator,
  title={Objectrelator: Enabling cross-view object relation understanding across ego-centric and exo-centric perspectives},
  author={Fu, Yuqian and Wang, Runze and Ren, Bin and Sun, Guolei and Gong, Biao and Fu, Yanwei and Paudel, Danda Pani and Huang, Xuanjing and Van Gool, Luc},
  booktitle={2025 IEEE/CVF International Conference on Computer Vision (ICCV)},
  pages={6530--6540},
  year={2025},
  organization={IEEE}
}

@inproceedings{kmeans,
  author    = {Sculley, D.},
  title     = {Web-Scale K-Means Clustering},
  booktitle = {Proceedings of the 19th International Conference on World Wide Web},
  pages     = {1177--1178},
  year      = {2010},
  doi       = {10.1145/1772690.1772862}
}

@misc{prism,
      title={PRISM: Self-Pruning Intrinsic Selection Method for Training-Free Multimodal Data Selection}, 
      author={Jinhe Bi and Aniri and Zengjie Jin and Yifan Wang and Danqi Yan and Wenke Huang and Xiaowen Ma and Sikuan Yan and Artur Hecker and Mang Ye and Xun Xiao and Hinrich Schuetze and Volker Tresp and Yunpu Ma},
      year={2026},
      eprint={2502.12119},
      archivePrefix={arXiv},
      primaryClass={cs.CV},
      url={https://arxiv.org/abs/2502.12119}, 
}

@misc{scalselect,
      title={ScalSelect: Scalable Training-Free Multimodal Data Selection for Efficient Visual Instruction Tuning}, 
      author={Changti Wu and Jiahuai Mao and Yuzhuo Miao and Shijie Lian and Bin Yu and Xiaopeng Lin and Cong Huang and Lei Zhang and Kai Chen},
      year={2026},
      eprint={2602.11636},
      archivePrefix={arXiv},
      primaryClass={cs.CV},
      url={https://arxiv.org/abs/2602.11636}, 
}

@misc{rds,
      title={Large-Scale Data Selection for Instruction Tuning}, 
      author={Hamish Ivison and Muru Zhang and Faeze Brahman and Pang Wei Koh and Pradeep Dasigi},
      year={2025},
      eprint={2503.01807},
      archivePrefix={arXiv},
      primaryClass={cs.CL},
      url={https://arxiv.org/abs/2503.01807}, 
}

@misc{Pistachio,
      title={Pistachio: Towards Synthetic, Balanced, and Long-Form Video Anomaly Benchmarks}, 
      author={Jie Li and Hongyi Cai and Mingkang Dong and Muxin Pu and Shan You and Fei Wang and Tao Huang},
      year={2026},
      eprint={2511.19474},
      archivePrefix={arXiv},
      primaryClass={cs.CV},
      url={https://arxiv.org/abs/2511.19474}, 
}

@misc{Evo-Depth,
      title={Evo-Depth: A Lightweight Depth-Enhanced Vision-Language-Action Model}, 
      author={Tao Lin and Yuxin Du and Jiting Liu and Nuobei Zhu and Yunhe Li and Yuqian Fu and Yinxinyu Chen and Hongyi Cai and Zewei Ye and Bing Cheng and Kai Ye and Yiran Mao and Yilei Zhong and MingKang Dong and Junchi Yan and Gen Li and Bo Zhao},
      year={2026},
      eprint={2605.14950},
      archivePrefix={arXiv},
      primaryClass={cs.CV},
      url={https://arxiv.org/abs/2605.14950}, 
}

@misc{Autodebias,
      title={AutoDebias: Automated Framework for Debiasing Text-to-Image Models}, 
      author={Hongyi Cai and Mohammad Mahdinur Rahman and Mingkang Dong and Muxin Pu and Moqyad Alqaily and Jie Li and Xinfeng Li and Jialie Shen and Meikang Qiu and Qingsong Wen},
      year={2026},
      eprint={2508.00445},
      archivePrefix={arXiv},
      primaryClass={cs.CV},
      url={https://arxiv.org/abs/2508.00445}, 
}

@misc{ObjectStream,
      title={ObjectStream: Latent Objects as Memory Anchors for Streaming Video Understanding}, 
      author={Mingkang Dong and Muxin Pu and Jie Li and Bohan Guo and Songruo Chen and Bin Ren and Xu Zheng and Chen Zhao and Tianwen Qian and Mohamed Elhoseiny and Yuqian Fu},
      year={2026},
      eprint={2607.28312},
      archivePrefix={arXiv},
      primaryClass={cs.CV},
      url={https://arxiv.org/abs/2607.28312}, 
}

@inproceedings{hu2022lora,
  title     = {LoRA: Low-Rank Adaptation of Large Language Models},
  author    = {Hu, Edward J. and Shen, Yelong and Wallis, Phillip and
               Allen-Zhu, Zeyuan and Li, Yuanzhi and Wang, Shean and
               Wang, Lu and Chen, Weizhu},
  booktitle = {International Conference on Learning Representations (ICLR)},
  year      = {2022}
}

@inproceedings{lee2024coincide,
      title={Concept-skill Transferability-based Data Selection for Large Vision-Language Models},
      author={Jaewoo Lee and Boyang Li and Sung Ju Hwang},
      year={2024},
      booktitle={Proceedings of the Conference on Empirical Methods in Natural Language Processing (EMNLP)},
}

@misc{yan2025coidoefficientdataselection,
      title={CoIDO: Efficient Data Selection for Visual Instruction Tuning via Coupled Importance-Diversity Optimization}, 
      author={Yichen Yan and Ming Zhong and Qi Zhu and Xiaoling Gu and Jinpeng Chen and Huan Li},
      year={2025},
      eprint={2510.17847},
      archivePrefix={arXiv},
      primaryClass={cs.CV},
      url={https://arxiv.org/abs/2510.17847}, 
}

@inproceedings{clipscore,
  title={Learning transferable visual models from natural language supervision},
  author={Radford, Alec and Kim, Jong Wook and Hallacy, Chris and Ramesh, Aditya and Goh, Gabriel and Agarwal, Sandhini and Sastry, Girish and Askell, Amanda and Mishkin, Pamela and Clark, Jack and Krueger, Gretchen and Sutskever, Ilya},
  booktitle={International Conference on Machine Learning (ICML)},
  year={2021}
}

@inproceedings{chen-etal-2024-vision,
    title = "Your Vision-Language Model Itself Is a Strong Filter: Towards High-Quality Instruction Tuning with Data Selection",
    author = "Chen, Ruibo  and
      Wu, Yihan  and
      Chen, Lichang  and
      Liu, Guodong  and
      He, Qi  and
      Xiong, Tianyi  and
      Liu, Chenxi  and
      Guo, Junfeng  and
      Huang, Heng",
    editor = "Ku, Lun-Wei  and
      Martins, Andre  and
      Srikumar, Vivek",
    booktitle = "Findings of the Association for Computational Linguistics: ACL 2024",
    month = aug,
    year = "2024",
    address = "Bangkok, Thailand",
    publisher = "Association for Computational Linguistics",
    url = "https://aclanthology.org/2024.findings-acl.246/",
    doi = "10.18653/v1/2024.findings-acl.246",
    pages = "4156--4172"
}

@inproceedings{sqa,
  title={ScienceQA: A Benchmark for Science Question Answering through Logical Reasoning},
  author={Lu, Pan and Mishra, Swaroop and Xia, Tony and Qiu, Liang and Chang, Kai-Wei and Zhu, Song-Chun and Majumder, Oyvind and Gopi, Aditya},
  booktitle={NeurIPS},
  year={2022}
}

@inproceedings{pope,
  title={Evaluating Object Hallucination in Large Vision-Language Models},
  author={Li, Yifan and Du, Yifan and Zhou, Kun and Wang, Jinpeng and Zhao, Wayne Xin and Wen, Ji-Rong},
  booktitle={EMNLP},
  year={2023}
}

@misc{li2024quantityqualityboostingllm,
      title={From Quantity to Quality: Boosting LLM Performance with Self-Guided Data Selection for Instruction Tuning}, 
      author={Ming Li and Yong Zhang and Zhitao Li and Jiuhai Chen and Lichang Chen and Ning Cheng and Jianzong Wang and Tianyi Zhou and Jing Xiao},
      year={2024},
      eprint={2308.12032},
      archivePrefix={arXiv},
      primaryClass={cs.CL},
      url={https://arxiv.org/abs/2308.12032}, 
}

@inproceedings{llava,
  title={Visual instruction tuning},
  author={Liu, Haotian and Li, Chunyuan and Wu, Qingyang and Lee, Yong Jae},
  booktitle={NeurIPS},
  year={2024}
}

@article{liu2023mmbench,
  title={Mmbench: Is your multi-modal model an all-around player?},
  author={Liu, Yuan and Duan, Haodong and Zhang, Yuanhan and others},
  journal={arXiv preprint arXiv:2307.06281},
  year={2023}
}

@misc{kdpp,
      title={Efficient Sampling for k-Determinantal Point Processes}, 
      author={Chengtao Li and Stefanie Jegelka and Suvrit Sra},
      year={2016},
      eprint={1509.01618},
      archivePrefix={arXiv},
      primaryClass={cs.LG},
      url={https://arxiv.org/abs/1509.01618}, 
}

@misc{perplexity,
      title={Example Perplexity}, 
      author={Nevin L. Zhang and Weiyan Xie and Zhi Lin and Guanfang Dong and Xiao-Hui Li and Caleb Chen Cao and Yunpeng Wang},
      year={2022},
      eprint={2203.08813},
      archivePrefix={arXiv},
      primaryClass={cs.LG},
      url={https://arxiv.org/abs/2203.08813}, 
}

@misc{ai2d,
      title={A Diagram Is Worth A Dozen Images}, 
      author={Aniruddha Kembhavi and Mike Salvato and Eric Kolve and Minjoon Seo and Hannaneh Hajishirzi and Ali Farhadi},
      year={2016},
      eprint={1603.07396},
      archivePrefix={arXiv},
      primaryClass={cs.CV},
      url={https://arxiv.org/abs/1603.07396}, 
}

@misc{dose,
      title={DOSE: Data Selection for Multi-Modal LLMs via Off-the-Shelf Models}, 
      author={Biao Wu and Yiwu Zhong and Meng Fang and Ling Chen},
      year={2026},
      eprint={2604.16979},
      archivePrefix={arXiv},
      primaryClass={cs.CV},
      url={https://arxiv.org/abs/2604.16979}, 
}

@misc{dics,
      title={DICS: Exploring Data Intrinsic Consistency for Visual Instruction Selection}, 
      author={Yuyang Hong and Jinhui Guo and Jiaqi Gu and Lubin Fan and Ruixiang Wang and Kun Ding and Yue Wu and Shiming Xiang and Jieping Ye},
      year={2026},
      eprint={2608.30209},
      archivePrefix={arXiv},
      primaryClass={cs.CV},
      url={https://arxiv.org/abs/2608.30209}, 
}

@article{ocrbench,
   title={OCRBench: on the hidden mystery of OCR in large multimodal models},
   volume={67},
   ISSN={1869-1919},
   url={http://dx.doi.org/10.1007/s11432-024-4235-6},
   DOI={10.1007/s11432-024-4235-6},
   number={12},
   journal={Science China Information Sciences},
   publisher={Springer Science and Business Media LLC},
   author={Liu, Yuliang and Li, Zhang and Huang, Mingxin and Yang, Biao and Yu, Wenwen and Li, Chunyuan and Yin, Xu-Cheng and Liu, Cheng-Lin and Jin, Lianwen and Bai, Xiang},
   year={2024},
   month=Dec }

@inproceedings{li2026egocross,
  title={Egocross: Benchmarking multimodal large language models for cross-domain egocentric video question answering},
  author={Li, Yanjun and Fu, Yuqian and Qian, Tianwen and Xu, Qi'ao and Dai, Silong and Paudel, Danda Pani and Van Gool, Luc and Wang, Xiaoling},
  booktitle={Proceedings of the AAAI Conference on Artificial Intelligence},
  volume={40},
  number={8},
  pages={6592--6600},
  year={2026}
}

@inproceedings{balauca2025understanding,
  title={Understanding Museum Exhibits using Vision-Language Reasoning},
  author={Balauca, Ada-Astrid and Garai, Sanjana and Balauca, Stefan and Shetty, Rasesh Udayakumar and Agrawal, Naitik and Shah, Dhwanil Subhashbhai and Fu, Yuqian and Wang, Xi and Toutanova, Kristina and Paudel, Danda Pani and others},
  booktitle={2025 IEEE/CVF International Conference on Computer Vision (ICCV)},
  pages={2227--2238},
  year={2025},
  organization={IEEE}
}

@inproceedings{pan2026v,
  title={V\^{} 2-SAM: Marrying SAM2 with Multi-Prompt Experts for Cross-View Object Correspondence},
  author={Pan, Jiancheng and Wang, Runze and Qian, Tianwen and Mahdi, Mohammad and Fu, Yanwei and Xue, Xiangyang and Huang, Xiaomeng and Van Gool, Luc and Paudel, Danda Pani and Fu, Yuqian},
  booktitle={Proceedings of the IEEE/CVF Conference on Computer Vision and Pattern Recognition},
  pages={16910--16919},
  year={2026}
}

@misc{XMAS,
      title={Data Selection for Fine-tuning Vision Language Models via Cross Modal Alignment Trajectories}, 
      author={Nilay Naharas and Dang Nguyen and Nesihan Bulut and Mohammadhossein Bateni and Vahab Mirrokni and Baharan Mirzasoleiman},
      year={2025},
      eprint={2510.01454},
      archivePrefix={arXiv},
      primaryClass={cs.CV},
      url={https://arxiv.org/abs/2510.01454}, 
}

@misc{MME,
      title={MME: A Comprehensive Evaluation Benchmark for Multimodal Large Language Models}, 
      author={Chaoyou Fu and Peixian Chen and Yunhang Shen and Yulei Qin and Mengdan Zhang and Xu Lin and Jinrui Yang and Xiawu Zheng and Ke Li and Xing Sun and Yunsheng Wu and Rongrong Ji and Caifeng Shan and Ran He},
      year={2025},
      eprint={2306.13394},
      archivePrefix={arXiv},
      primaryClass={cs.CV},
      url={https://arxiv.org/abs/2306.13394}, 
}

@misc{icons,
      title={ICONS: Influence Consensus for Vision-Language Data Selection}, 
      author={Xindi Wu and Mengzhou Xia and Rulin Shao and Zhiwei Deng and Pang Wei Koh and Olga Russakovsky},
      year={2025},
      eprint={2501.00654},
      archivePrefix={arXiv},
      primaryClass={cs.CV},
      url={https://arxiv.org/abs/2501.00654}, 
}

@misc{qwen3,
      title={Qwen3-VL Technical Report}, 
      author={Shuai Bai and Yuxuan Cai and Ruizhe Chen and Keqin Chen and Xionghui Chen and Zesen Cheng and Lianghao Deng and Wei Ding and Chang Gao and Chunjiang Ge and Wenbin Ge and Zhifang Guo and Qidong Huang and Jie Huang and Fei Huang and Binyuan Hui and Shutong Jiang and Zhaohai Li and Mingsheng Li and Mei Li and Kaixin Li and Zicheng Lin and Junyang Lin and Xuejing Liu and Jiawei Liu and Chenglong Liu and Yang Liu and Dayiheng Liu and Shixuan Liu and Dunjie Lu and Ruilin Luo and Chenxu Lv and Rui Men and Lingchen Meng and Xuancheng Ren and Xingzhang Ren and Sibo Song and Yuchong Sun and Jun Tang and Jianhong Tu and Jianqiang Wan and Peng Wang and Pengfei Wang and Qiuyue Wang and Yuxuan Wang and Tianbao Xie and Yiheng Xu and Haiyang Xu and Jin Xu and Zhibo Yang and Mingkun Yang and Jianxin Yang and An Yang and Bowen Yu and Fei Zhang and Hang Zhang and Xi Zhang and Bo Zheng and Humen Zhong and Jingren Zhou and Fan Zhou and Jing Zhou and Yuanzhi Zhu and Ke Zhu},
      year={2025},
      eprint={2511.21631},
      archivePrefix={arXiv},
      primaryClass={cs.CV},
      url={https://arxiv.org/abs/2511.21631}, 
}

@inproceedings{VisionFlan,
    title = "Vision-Flan: Scaling Human-Labeled Tasks in Visual Instruction Tuning",
    author = "Xu, Zhiyang  and
      Feng, Chao  and
      Shao, Rulin  and
      Ashby, Trevor  and
      Shen, Ying  and
      Jin, Di  and
      Cheng, Yu  and
      Wang, Qifan  and
      Huang, Lifu",
    editor = "Ku, Lun-Wei  and
      Martins, Andre  and
      Srikumar, Vivek",
    booktitle = "Findings of the Association for Computational Linguistics: ACL 2024",
    month = aug,
    year = "2024",
    address = "Bangkok, Thailand",
    publisher = "Association for Computational Linguistics",
    url = "https://aclanthology.org/2024.findings-acl.905/",
    doi = "10.18653/v1/2024.findings-acl.905",
    pages = "15271--15342"
}

@article{LRV,
  title={Aligning Large Multi-Modal Model with Robust Instruction Tuning},
  author={Liu, Fuxiao and Lin, Kevin and Li, Linjie and Wang, Jianfeng and Yacoob, Yaser and Wang, Lijuan},
  journal={arXiv preprint arXiv:2306.14565},
  year={2023}
}
\section{Appendix}

\subsection{Datasets}
\label{app:datasets}

We evaluate data selection methods on three multimodal instruction-tuning
datasets with substantially different data distributions.

\paragraph{LLaVA-625K.}
LLaVA-625K is the multimodal portion of the LLaVA-v1.5 instruction-tuning
mixture~\citep{llava}. The original mixture contains both multimodal
and text-only conversations. We retain the 624,610 examples with valid image
inputs and refer to this candidate pool as LLaVA-625K. It covers a broad
range of tasks, including general visual question answering, detailed image
description, OCR, knowledge-based reasoning, and region-level understanding.
In our default setting, we select 15\% of this pool, corresponding to
93,692 examples.

\paragraph{Vision-Flan-186K.}
Vision-Flan~\citep{VisionFlan} contains 186,103 human-annotated
vision-language examples collected from 191 tasks. Compared with the
heterogeneous LLaVA mixture, Vision-Flan has a more explicitly
task-structured distribution and includes classification, visual reasoning,
grounding, and knowledge-intensive tasks. Under a 15\% budget, the selected
subset contains 27,916 examples.

\paragraph{LRV-Sub-180K.}
LRV-Sub is constructed from the LRV-Instruction
dataset~\citep{LRV}, which contains both positive and negative
visual instructions designed to improve multimodal robustness. After
removing ten records with empty responses, the resulting candidate pool
contains 180,712 valid image–instruction–response triples. We select
27,107 examples, corresponding to 15\% of the valid pool. The distribution
of LRV-Sub differs substantially from LLaVA-625K because of its
contrastive positive and negative instructions.
\begin{table}[h]
\centering
\caption{Summary of the instruction-tuning datasets. The selected sizes
correspond to the default 15\% budget.}
\label{tab:dataset_statistics}
\renewcommand{\arraystretch}{1.1}
\begin{tabular}{lrrl}
\toprule
\textbf{Dataset} & \textbf{Full Pool} & \textbf{Selected} &
\textbf{Primary Characteristics} \\
\midrule
LLaVA-625K    & 624,610 & 93,692 & Heterogeneous instruction mixture \\
Vision-Flan-186K & 186,103 & 27,916 & Human-labeled, task-structured data \\
LRV-Sub-180K    & 180,712 & 27,107 & Positive and negative instructions \\
\bottomrule
\end{tabular}
\end{table}

For all datasets, the original assistant response is preserved for downstream
fine-tuning but is not used by our selector. Sample selection depends only on
the image and user instruction, preventing the selector from exploiting
response-side annotations or downstream benchmark labels.

\subsection{Baselines}
\label{app:baselines}

We compare our approach with heuristic, representation-based, model-based,
and multimodal-specific data selection methods. Whenever possible, the
selected subsets are used to fine-tune the same target model under our
unified training configuration.

\paragraph{Full Data Training.}
The target MLLM is trained using the complete candidate dataset. This setting
serves as the full-data reference and is assigned a relative performance of
100\%.

\paragraph{Random.}
We uniformly sample examples without replacement using the same data budget
as the corresponding selection experiment. Random selection measures whether
a method provides benefits beyond reducing the number of training examples.

\paragraph{Length.}
Length assigns each example an importance score according to its instruction
and response length, retaining examples with longer textual supervision. This
baseline tests whether the observed improvements can be explained simply by
selecting samples containing more tokens.

\paragraph{Perplexity.}
Perplexity ranks examples according to their language-model perplexity. We
follow the commonly used middle-perplexity strategy, which removes both
trivial low-perplexity examples and extremely high-perplexity examples that
are more likely to be noisy.

\paragraph{COINCIDE.}
COINCIDE~\citep{lee2024coincide} extracts internal activations from a
lightweight reference model and clusters samples according to their
concept--skill composition. Its sampling strategy jointly considers cluster
density and cross-cluster transferability to preserve diverse and broadly
transferable training examples.

\paragraph{PRISM.}
PRISM~\citep{prism} is a training-free multimodal selection method that
uses the intrinsic representation geometry of a pretrained multimodal model.
It corrects representation degeneration and scores the contribution of
individual examples without training an additional selector or computing
per-sample gradients.

\paragraph{RDS+.}
RDS+~\citep{rds} is a scalable representation-based selection
method. It constructs instruction representations using weighted mean pooling
over pretrained language-model hidden states and selects a diverse subset in
the resulting semantic space. For multimodal data, we apply RDS+ to the
textual instruction associated with each image.

\paragraph{ScalSelect.}
ScalSelect~\citep{scalselect} extracts instruction-conditioned visual
representations from the target MLLM and estimates how strongly each example
contributes to the dominant representation subspace. It avoids pairwise
similarity computation and provides a training-free, linear-time selection
baseline.

\paragraph{VisNec.}
VisNec~\citep{VisNec} measures visual necessity through the difference
between the text-only loss and the multimodal loss of an instruction example.
Samples with large positive loss reductions require stronger visual evidence.
VisNec combines this score with semantic clustering to preserve task
diversity.

\paragraph{XMAS.}
XMAS~\citep{XMAS} trains a small proxy MLLM and records the trajectories of the
leading singular values of its cross-modal attention matrices. It clusters
examples according to these training trajectories and performs balanced
sampling across the resulting clusters. When included, XMAS is reproduced
under the same final data budget and target-model training protocol.

\paragraph{Fair-comparison protocol.}
For methods that select from the original candidate pool, we retain the
selected image, instruction, and response without rewriting them. We then
fine-tune the same model initialization using the same optimizer,
hyperparameters, random seed, and evaluation code. If a released method fixes
its subset to a slightly different size, such as 100K examples, we report its
actual sample count rather than describing it as an exact 15\% subset.

\subsection{Evaluation Benchmarks}
\label{app:evaluation_benchmarks}

We evaluate the fine-tuned models on nine complementary metrics covering
general multimodal understanding, perception, cognition, diagram reasoning,
hallucination, scientific question answering, and optical character
recognition.

\paragraph{MMBench.}
MMBench~\citep{liu2023mmbench} evaluates fine-grained multimodal perception
and reasoning with multiple-choice questions. We report both the English
split (MMBench-En) and Chinese split (MMBench-Cn). Following the strict
protocol, we use \emph{circular accuracy}: each question is evaluated under
multiple answer-option permutations and is counted as correct only according
to the circular scoring rule. These values should not be compared directly
with the less restrictive single-order or vanilla MMBench scores.

\paragraph{MME.}
MME~\citep{MME} evaluates multimodal capabilities using binary
yes/no questions. We separately report the aggregate score on the perception
subset (MME-P) and the cognition subset (MME-C). MME-P covers tasks such as
existence, count, position, color, and OCR, whereas MME-C evaluates
commonsense reasoning, numerical calculation, translation, and code
reasoning.

\paragraph{AI2D.}
AI2D~\citep{ai2d} measures diagram understanding and
diagram-grounded reasoning. We evaluate all 3,088 examples and report
exact-match multiple-choice accuracy.

\paragraph{POPE.}
POPE~\citep{pope} evaluates object hallucination through binary
questions constructed using random, popular, and adversarial negative
sampling. We report accuracy (POPE-A) and precision (POPE-P), each
macro-averaged over the three POPE subsets. Here, ``A'' and ``P'' denote
\emph{accuracy} and \emph{precision}, rather than adversarial and popular
splits.

\paragraph{ScienceQA-IMG.}
ScienceQA~\citep{sqa} evaluates scientific knowledge and multimodal
reasoning. We report accuracy on the image-dependent subset, denoted
SQA-IMG. This value is distinct from the accuracy over the complete ScienceQA
test set, which also contains questions that do not require an image.

\paragraph{OCRBench.}
OCRBench~\citep{ocrbench} contains 1,000 questions covering text
recognition, scene-text understanding, document-oriented VQA, handwritten
text recognition, and text-centric reasoning. The official raw score ranges
from 0 to 1,000. For consistency with the other percentage-scale metrics, the
tables report the raw score divided by ten.

All evaluations use deterministic decoding with temperature zero. The same
benchmark snapshots, prompts, conversation template, maximum generation
length, parsers, and scoring scripts are used for every compared method.

\noindent\textbf{Evaluation Protocol.}
For MMBench, we adopt the stricter circular evaluation protocol, under which
a question is counted as correct only when the model answers all option
rotations consistently. For POPE, POPE-A and POPE-P denote the macro-averaged
accuracy and precision over the three official subsets, respectively.
SQA-IMG includes only ScienceQA questions with image context. AI2D is evaluated
using exact multiple-choice accuracy. MME-P and MME-C are reported in their
original point scales, whereas OCRBench is normalized to a $0$--$100$ scale
for consistent presentation.\\

\subsection{LLaVA Fine-Tuning Details}
\label{app:llava_training}

For all LLaVA experiments, we initialize the language backbone from
Vicuna-7B-v1.5 and use the frozen CLIP ViT-L/14-336 vision tower. The
multimodal projector is initialized from the pretrained LLaVA-v1.5 projector
and consists of a two-layer MLP with a GELU activation
(\texttt{mlp2x\_gelu}), mapping the 1024-dimensional visual features to the
4096-dimensional language-model hidden space. We use visual features from the
penultimate CLIP layer and pad images to a square aspect ratio. The vision
tower remains frozen, whereas the multimodal projector is updated during
instruction tuning.

We fine-tune the language model using LoRA. The adapters are inserted into all
linear transformations of the self-attention and feed-forward modules,
corresponding to
\texttt{q\_proj}, \texttt{k\_proj}, \texttt{v\_proj},
\texttt{o\_proj}, \texttt{gate\_proj}, \texttt{up\_proj}, and
\texttt{down\_proj}. We set the LoRA rank to $r=128$, the scaling parameter to
$\alpha=256$, and the dropout rate to $0.05$. No bias parameters are adapted.
The original language-model parameters and the vision tower are frozen; only
the LoRA parameters and multimodal projector are optimized.

All models are trained for one epoch using AdamW. The learning rate is
$2\times10^{-4}$ for the LoRA parameters and $2\times10^{-5}$ for the
multimodal projector, with zero weight decay. We apply a cosine learning-rate
schedule with a warm-up ratio of $0.03$. The maximum sequence length is 2048,
and training uses BF16, TF32, gradient checkpointing, lazy preprocessing, and
modality-length grouping. The training seed is fixed to 17.
\begin{table}[h]
\centering
\caption{Fine-tuning configuration for LLaVA-v1.5-7B.}
\label{tab:llava_training_details}
\renewcommand{\arraystretch}{1.1}
\begin{tabular}{l|l}
\toprule
\textbf{Configuration} & \textbf{Value} \\
\midrule
Language backbone & Vicuna-7B-v1.5 \\
Vision tower & CLIP ViT-L/14-336, frozen \\
Visual feature layer & Penultimate layer ($-2$) \\
Multimodal projector & 2-layer MLP with GELU \\
LoRA rank / alpha & 128 / 256 \\
LoRA dropout / bias & 0.05 / none \\
LoRA learning rate & $2\times10^{-4}$ \\
Projector learning rate & $2\times10^{-5}$ \\
Optimizer & AdamW \\
Weight decay & 0 \\
Schedule & Cosine, 3\% warm-up \\
Training epochs & 1 \\
Per-device batch size & 16 \\
Effective global batch size & 128 \\
Maximum sequence length & 2048 \\
Numerical precision & BF16 and TF32 \\
Distributed optimization & DeepSpeed ZeRO Stage~3 \\
Training seed & 17 \\
\bottomrule
\end{tabular}
\end{table}
\subsection{Relative Performance}
\label{app:relative_performance}

To summarize performance across metrics with different numerical scales, we
report average relative performance. Let $s_m$ denote the score of a selected
subset on metric $m$, and let $s_m^{\mathrm{full}}$ denote the corresponding
full-data score. We compute

\begin{equation}
\mathrm{Rel.}
=
\frac{100}{M}
\sum_{m=1}^{M}
\frac{s_m}{s_m^{\mathrm{full}}},
\label{eq:relative_performance}
\end{equation}

where $M=9$ corresponds to MMBench-En, MMBench-Cn, MME-P, MME-C,
AI2D, POPE-A, POPE-P, SQA-IMG, and OCRBench. Each metric contributes
equally regardless of its original scale. The full-data reference therefore
has $\mathrm{Rel.}=100\%$. Relative performance is computed within each
dataset or model block using its corresponding full-data reference.

\vspace{-10pt}
\subsection{Selector Implementation Details}
\label{app:selector_details}
\vspace{-10pt}
Our default selector uses a frozen CLIP ViT-B/32
encoder~\citep{clipscore}. Image and instruction embeddings are
separately $\ell_2$-normalized and concatenated to construct the multimodal
representation. We apply MiniBatch K-Means with $K=128$ and seed 42.

Within each cluster, the centroid-nearest 3\% of the examples form the
pseudo-labeled core. We reserve 20\% of each cluster-specific core for
validation and use the remaining 80\% for selector training. Thus, the 20\%
validation ratio applies only to the small core set, not to the complete
candidate dataset. Inverse-class-frequency sampling prevents large clusters
from dominating selector optimization.

The selector contains separate image and text projection branches that map
both modalities into a 256-dimensional hidden space. A learned sigmoid gate
adaptively balances the two branches, while an additional interaction branch
models elementwise image--text correspondence. The fused representation is
passed through normalization, nonlinear activation, dropout, and a
$K$-class prediction head.

We train only the lightweight selector  using AdamW with a
learning rate of $1\times10^{-5}$; the CLIP encoder remains frozen. At
selection time, the selector is also frozen. We construct a candidate pool
from the lowest-confidence examples within each assigned semantic cluster, with a
pool size twice the assigned cluster quota. We then apply deterministic
blockwise $k$-DPP with a maximum block size of 128 to balance confidence-based
informativeness and multimodal diversity. Unless otherwise stated, the final
selection ratio is 15\%.

\subsection{Selector Checkpoint Selection}
\label{app:checkpoint}
To preserve residual uncertainty after the selector has learned the dominant
semantic structure, we select an intermediate checkpoint according to its
performance on the pseudo-label validation set. Let
\begin{equation}
A_{\mathrm{macro}}^{(e)}
=
\frac{1}{K}
\sum_{k=1}^{K}
\frac{n_{k,\mathrm{correct}}^{(e)}}{n_k}
\end{equation}
denote the validation macro accuracy at epoch $e$. Given a predefined target
$a_{\mathrm{target}}=0.60$, we select
\begin{equation}
e^*
=
\arg\min_{e\in\{1,\ldots,E\}}
\left|
A_{\mathrm{macro}}^{(e)}
-
a_{\mathrm{target}}
\right|.
\end{equation}
The selector is trained for $E=100$ epochs, and checkpoint selection depends
only on the cluster-pseudo-labeled validation split. For the default CLIP
ViT-B/32 selector, this criterion selects epoch 35, which achieves a validation
macro accuracy of $0.6108$ and a micro accuracy of $0.5807$, with predictions
covering 121 of the 128 semantic clusters. This frozen checkpoint is reused
for selection on all evaluated target datasets without retraining.
\begin{table}[h]
\centering
\caption{Selection of the default intermediate selector checkpoint.}
\label{tab:selector_checkpoint}
\small
\begin{tabular}{lc}
\toprule
\textbf{Setting} & \textbf{Value} \\
\midrule
Selector backbone & CLIP ViT-B/32 \\
Number of clusters & 128 \\
Target macro accuracy & 0.6000 \\
Selected epoch & 35 \\
Validation macro accuracy & 0.6108 \\
Validation micro accuracy & 0.5807 \\
Predicted-cluster coverage & 121/128 \\
\bottomrule
\end{tabular}
\end{table}

\noindent\textbf{Analysis of Selector Backbone}
Table~\ref{tab:selector_backbone} compares four representation backbones used
by the selector. Relative performance ranges from 98.6\% to 99.2\% across the
evaluated encoders. CLIP ViT-B/32 gives the highest aggregate result of 99.2\%,
while SigLIP2 and MetaCLIP lead individual metrics such as MMBench-En, AI2D,
and MME-P. We therefore use CLIP ViT-B/32 as the default because it provides
the best observed aggregate performance while requiring lower
feature-extraction and selection costs than the larger encoders.
\begin{table*}[h]
\centering
\caption{Effect of the representation backbone used by the data selector.
All selector variants retain 15\% (93.7K) of LLaVA-625K. The best result is \textbf{boled}.}
\label{tab:selector_backbone}
\renewcommand{\arraystretch}{1.15}
\resizebox{\textwidth}{!}{%
\begin{tabular}{
l|c|
c c c c c c c c c|c
}
\toprule
\textbf{Selector Backbone}
& \textbf{\#Samples}
& \textbf{MMBench-En}
& \textbf{MMBench-Cn}
& \textbf{MME-P}
& \textbf{MME-C}
& \textbf{AI2D}
& \textbf{POPE-A}
& \textbf{POPE-P}
& \textbf{SQA-IMG}
& \textbf{OCRBench}
& \textbf{Rel.} \\
\midrule

Full Data Training
& 625K
& 64.0
& 56.7
& 1463.9
& 278.6
& 53.8
& 85.2
& 93.1
& 67.6
& 31.3
& 100.0\% \\
\midrule
\rowcolor{blue!10}
CLIP ViT-B/32
& 93.7K
& 60.0
& 53.4
& 1420.6
& \textbf{314.6}
& 53.7
& \textbf{86.9}
& \textbf{92.9}
& 63.9
& \textbf{30.9}
& \textbf{99.2\%} \\

CLIP ViT-L/14-336
& 93.7K
& 60.2
& \textbf{54.1}
& 1436.2
& 293.6
& 53.1
& \textbf{86.9}
& 92.5
& \textbf{65.9}
& 30.4
& 98.6\% \\

SigLIP2 Base/16-384
& 93.7K
& \textbf{61.7}
& 53.7
& 1449.8
& 296.1
& \textbf{54.2}
& 86.6
& 92.3
& 63.7
& 30.6
& 98.9\% \\

MetaCLIP ViT-H/14
& 93.7K
& 61.0
& 53.3
& \textbf{1456.2}
& 301.5
& 54.1
& 86.8
& 92.6
& 63.0
& 30.1
& 98.7\% \\

\bottomrule
\end{tabular}}
\end{table*}

\subsection{Implementation Details of Diversity-Aware Sampling}
\label{app:kdpp}

After obtaining the selector confidence and assigned semantic cluster for each sample, we apply a cluster-wise $k$-DPP procedure to jointly preserve uncertainty and diversity. Let $N$ denote the number of candidate samples and let $M=\lceil\rho N\rceil$ be the target subset size, where $\rho=0.15$ by default. We first allocate a quota $m_c$ to each semantic cluster $c$ in proportion to its population, subject to
\[
\sum_{c=1}^{C}m_c=M,\qquad 0\leq m_c\leq n_c,
\]
where $n_c$ is the number of samples assigned to cluster $c$. Integer rounding is resolved using the largest-remainder rule, ensuring that the final subset contains exactly $M$ samples.

\paragraph{Low-confidence candidate pool.}
Within each cluster, samples are sorted in ascending order of selector confidence. Rather than directly selecting the lowest-confidence samples, we construct a candidate pool containing
\[
p_c=\min\!\left(n_c,\left\lceil\alpha m_c\right\rceil\right)
\]
samples, where $\alpha=2$ is the candidate multiplier. Thus, approximately the lowest-confidence 30\% of the complete dataset enters the diversity-selection stage when the final budget is 15\%. For a candidate with confidence rank $r_i\in\{0,\ldots,p_c-1\}$, we assign the rank-based quality weight
\[
q_i=1-\frac{r_i}{p_c}.
\]
Lower-confidence samples therefore receive larger quality weights without relying on the absolute calibration of selector probabilities.

\paragraph{Quality-aware $k$-DPP kernel.}
For each candidate, we concatenate its visual and textual CLIP representations and apply $\ell_2$ normalization:
\[
\mathbf{z}_i=
\frac{[\mathbf{v}_i;\mathbf{t}_i]}
{\|[\mathbf{v}_i;\mathbf{t}_i]\|_2}.
\]
The similarity between two candidates is defined using shifted cosine similarity,
\[
S_{ij}=\frac{1+\mathbf{z}_i^\top\mathbf{z}_j}{2}.
\]
We then construct the quality-weighted $L$-ensemble kernel
\[
L_{ij}=\sqrt{q_iq_j}\,S_{ij}+\epsilon\mathbbm{1}[i=j],
\]
where $\epsilon$ is a small diagonal jitter for numerical stability. The shifted cosine matrix is positive semidefinite because it is the average of a Gram matrix and an all-ones matrix. For a subset $Y$ of fixed size $k$, the sampling probability is
\[
P(Y\mid |Y|=k)
=
\frac{\det(L_Y)}
{\sum_{|Y'|=k}\det(L_{Y'})}.
\]
The quality terms favor uncertain samples, while the determinant discourages selecting candidates with redundant multimodal representations.

\paragraph{Blockwise exact sampling.}
Constructing a global $k$-DPP kernel is impractical at the scale of hundreds of thousands of samples. We therefore randomly permute each cluster-specific candidate pool using a fixed seed and divide it into blocks of at most $B=128$ samples. The cluster quota is distributed proportionally across these blocks, and exact $k$-DPP sampling is performed independently within every block. The union of the block-level samples forms the selected subset for that cluster. This procedure preserves the exact global selection budget while reducing the peak kernel memory from $\mathcal{O}(N^2)$ to $\mathcal{O}(B^2)$ and the eigendecomposition cost to approximately $\mathcal{O}(NB^2)$.

\subsection{Transfer Protocols and Sensitivity Analysis}
\label{app:transfer_protocols}

\paragraph{Transfer to unseen datasets.}
The selector used for Vision-Flan and LRV-Sub is trained only on
LLaVA-625K. We directly apply the frozen source selector to the target
dataset without target-specific pseudo-label construction, selector
fine-tuning, or downstream benchmark supervision. Target examples are
assigned to the semantic structure learned from LLaVA-625K, after which
cluster-wise low-confidence pooling and $k$-DPP selection are performed under
the same 15\% budget.

\paragraph{Transfer across model architectures.}
For the Qwen3-VL-4B and Qwen3-VL-8B experiments, we reuse the subset selected
by the default LLaVA-trained CLIP ViT-B/32 selector. We do not retrain the
selector or reselect the data for each target architecture. Consequently,
performance improvements on these models measure whether the selected data
are transferable across MLLM families rather than optimized specifically for
LLaVA-v1.5-7B.

\noindent\textbf{Analysis of the Selection Ratio.}
We evaluate selection ratios from 5\% to 25\%. As shown in
Tab.~\ref{tab:sensitivity_ratio}, OnceSelect attains 95.0\% relative
performance using 5\% of LLaVA-625K and reaches the highest observed value of
99.2\% at the 15\% budget. The 20\% and 25\% subsets obtain 98.2\% and 98.9\%,
respectively, showing that retaining more examples does not monotonically
improve aggregate performance. We therefore use 15\% as the default budget
based on the evaluated settings. This comparison identifies the best observed
ratio in our experiments rather than a universal optimum.

\begin{table*}[h]
\centering
\caption{Performance under different selection budgets. Our default ratio is 15\%. The best result is \textbf{bolded}.}
\label{tab:sensitivity_ratio}
\renewcommand{\arraystretch}{1.15}
\resizebox{\textwidth}{!}{%
\begin{tabular}{
l|c|
c c c c c c c c c|c
}
\toprule
\textbf{Ratio}
& \textbf{\#Samples}
& \textbf{MMBench-En}
& \textbf{MMBench-Cn}
& \textbf{MME-P}
& \textbf{MME-C}
& \textbf{AI2D}
& \textbf{POPE-A}
& \textbf{POPE-P}
& \textbf{SQA-IMG}
& \textbf{OCRBench}
& \textbf{Rel.} \\
\midrule

Full Data Training
& 625K
& 64.0
& 56.7
& 1463.9
& 278.6
& 53.8
& 85.2
& 93.1
& 67.6
& 31.3
& 100.0\% \\
\midrule
5\%
& 31.2K
& 55.6
& 51.1
& 1285.2
& 300.1
& 50.9
& 86.4
& 89.0
& 63.9
& 30.0
& 95.0\%\\

10\%
& 62.5K
& 60.2
& 52.9
& 1388.1
& 282.1
& 53.0
& 86.6
& 92.8
& \textbf{66.1}
& 30.3
& 97.5\% \\

\rowcolor{blue!10}
15\%
& 93.7K
& 60.0
& 53.4
& 1420.6
& \textbf{314.6}
& 53.7
& \textbf{86.9}
& \textbf{92.9}
& 63.9
& \textbf{30.9}
& \textbf{99.2\%} \\

20\%
& 124.9K
& 61.0
& 54.5
& 1407.9
& 293.6
& 52.7
& 87.0
& 92.5
& 64.3
& 30.1
& 98.2\% \\

25\%
& 156.2K
& 62.5
& 54.6
& 1420.9
& 297.5
& 53.4
& 86.7
& 92.6
& 63.9
& 30.5
& 98.9\%\\
\bottomrule
\end{tabular}}
\end{table*}
\noindent\textbf{Analysis of the Number of Clusters.}
The number of clusters $K$ controls the granularity of the semantic partition.
At a fixed 15\% budget, relative performance ranges from 97.2\% to 99.2\% over
$K\in\{32,64,128,256\}$, with the highest observed value at $K=128$. This
pattern is consistent with a granularity trade-off: a coarse partition may
combine heterogeneous examples, whereas a very fine partition provides fewer
examples for estimating each cluster-specific confidence signal. We therefore
use $K=128$ in the main experiments.
\begin{table*}[h]
\centering
\caption{Sensitivity to the number of semantic clusters $K$ on LLaVA-625K.
All selection variants retain 15\% (93.7K) of the training data. The best result is \textbf{bolded}.}
\label{tab:k_sensitivity}
\renewcommand{\arraystretch}{1.15}
\resizebox{\textwidth}{!}{%
\begin{tabular}{
l|c|
c c c c c c c c c|c
}
\toprule
\textbf{Setting}
& \textbf{\#Samples}
& \textbf{MMBench-En}
& \textbf{MMBench-Cn}
& \textbf{MME-P}
& \textbf{MME-C}
& \textbf{AI2D}
& \textbf{POPE-A}
& \textbf{POPE-P}
& \textbf{SQA-IMG}
& \textbf{OCRBench}
& \textbf{Rel.} \\
\midrule

Full Data
& 625K
& 64.0
& 56.7
& 1463.9
& 278.6
& 53.8
& 85.2
& 93.1
& 67.6
& 31.3
& 100.0\% \\
\midrule

$K=32$
& 93.7K
& \textbf{61.4}
& \textbf{53.8}
& \textbf{1428.9}
& 302.9
& 52.6
& \textbf{86.3}
& 91.9
& 61.6
& \textbf{30.7}
& \textbf{98.2\%} \\

$K=64$
& 93.7K
& 59.8
& \textbf{53.8}
& 1411.9
& 293.6
& 53.2
& 85.5
& \textbf{93.5}
& \textbf{64.2}
& 28.6
& 97.3\% \\

\rowcolor{blue!10}
$K=128$ (Default)
& 93.7K
& 60.0
& 53.4
& 1420.6
& \textbf{314.6}
& \textbf{53.7}
& \textbf{86.9}
& 92.9
& 63.9
& \textbf{30.9}
& \textbf{99.2\%} \\

$K=256$
& 93.7K
& 59.3
& 52.6
& 1423.8
& 300.1
& 52.3
& 85.8
& 90.7
& 63.3
& 29.9
& 97.2\% \\

\bottomrule
\end{tabular}}
\end{table*}

\subsection{Visualization}
\label{sec:visualization}

\begin{figure*}[h]
    \centering

    \begin{subfigure}[t]{0.32\textwidth}
        \centering
        \includegraphics[width=\linewidth]{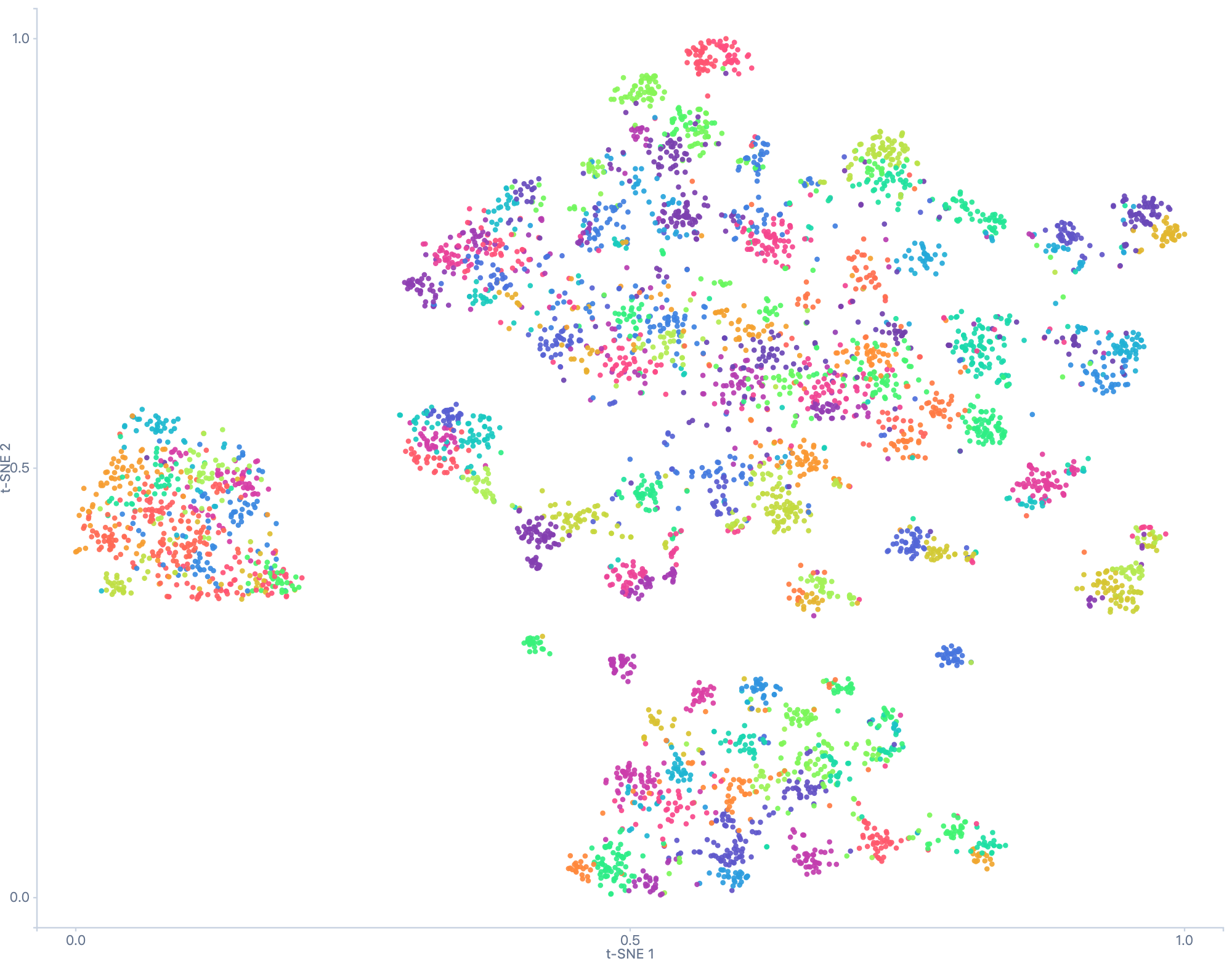}
        \caption{Complete source dataset}
        \label{fig:source_complete}
    \end{subfigure}
    \hfill
    \begin{subfigure}[t]{0.32\textwidth}
        \centering
        \includegraphics[width=\linewidth]{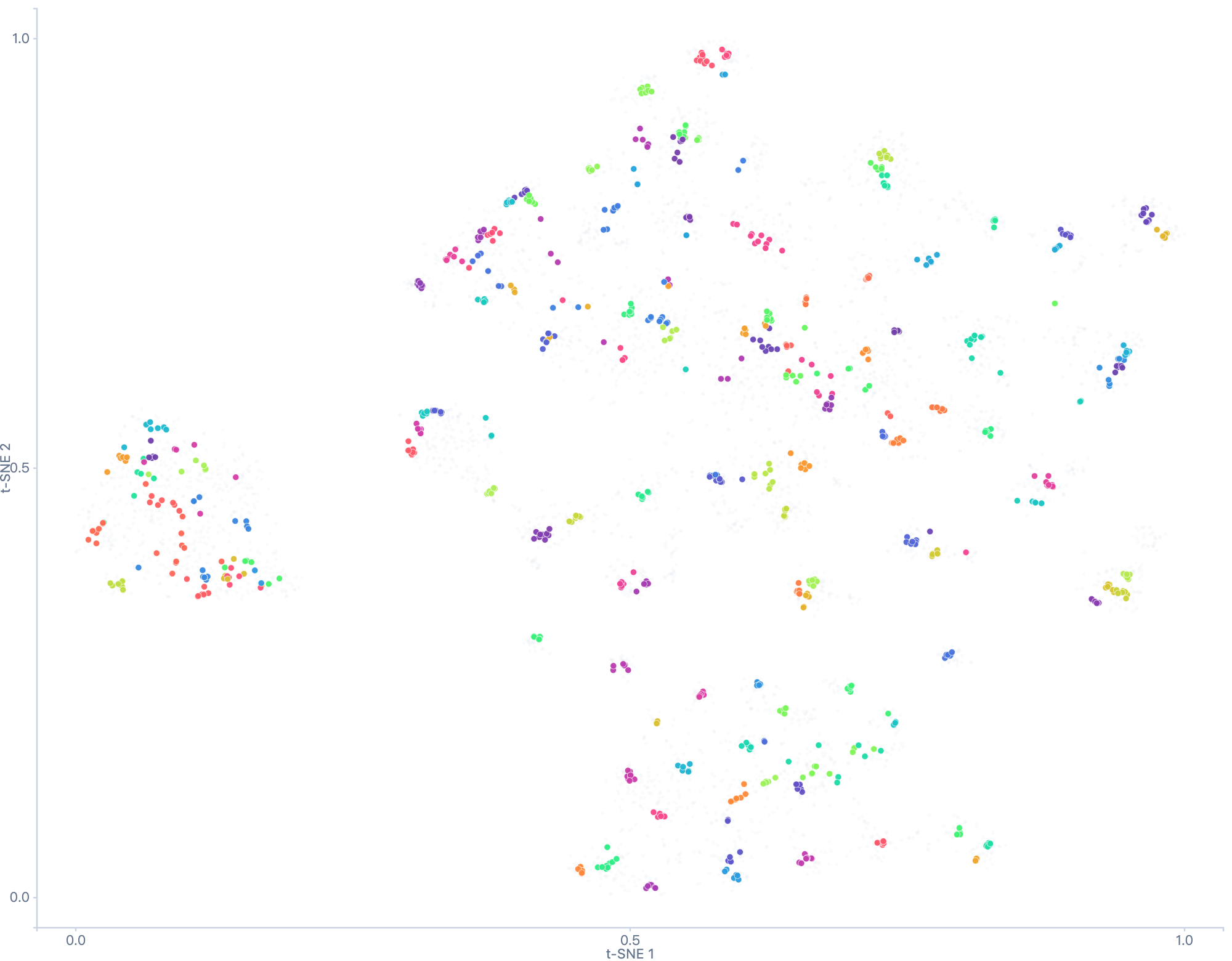}
        \caption{Pseudo-labeled core}
        \label{fig:source_core}
    \end{subfigure}
    \hfill
    \begin{subfigure}[t]{0.32\textwidth}
        \centering
        \includegraphics[width=\linewidth]{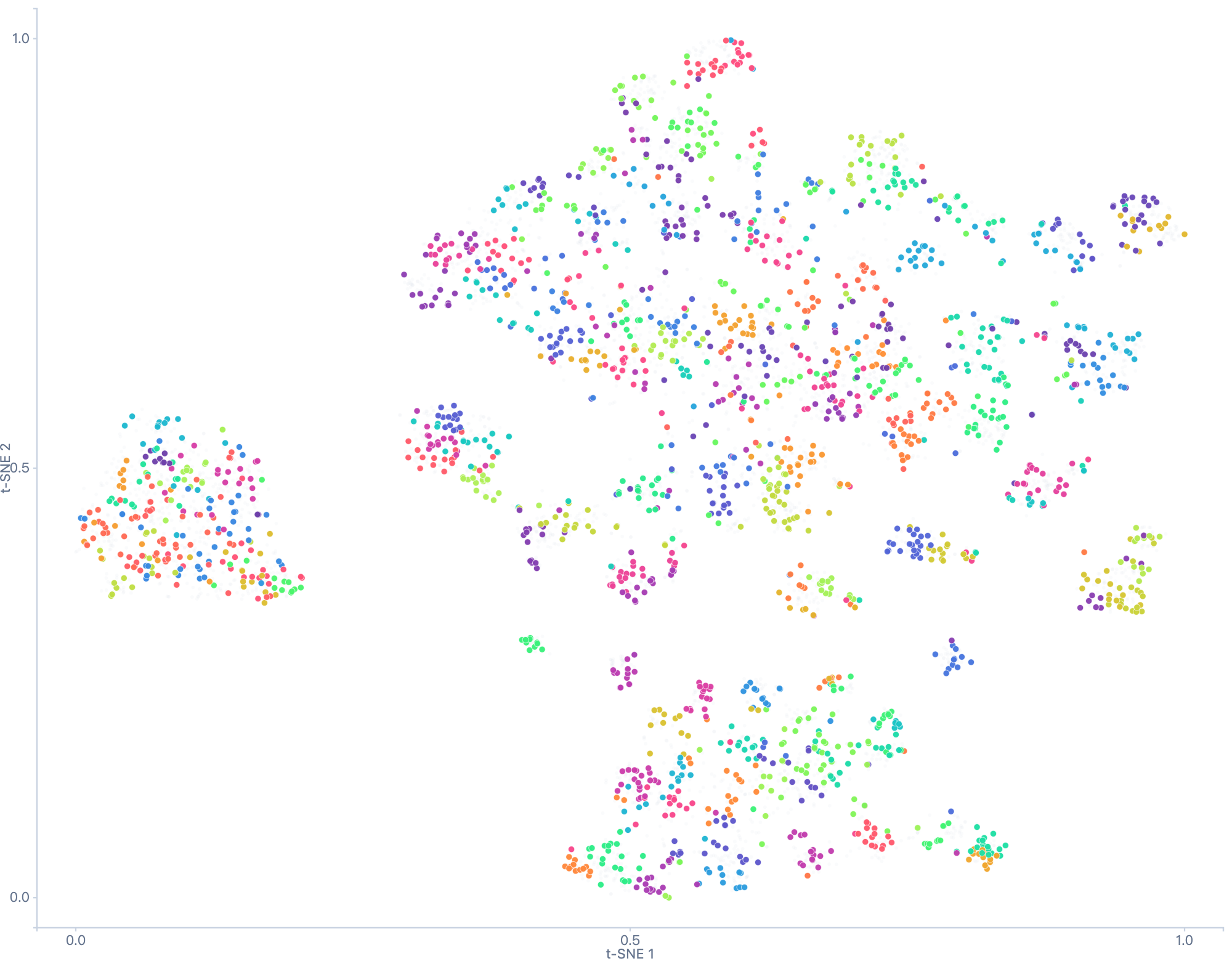}
        \caption{Final 15\% subset}
        \label{fig:source_selected}
    \end{subfigure}

    \caption{Visualization of the OnceSelect pipeline on LLaVA-625K:
    the complete source dataset, the centroid-nearest core used for selector
    training, and the final subset selected using uncertainty and $k$-DPP.}
    \label{fig:main_analysis}
\end{figure*}
\noindent\textbf{Visualization of the Selection Process.}
Fig.~\ref{fig:main_analysis} visualizes the LLaVA-625K representations
extracted by frozen CLIP ViT-B/32 and projected using a shared t-SNE mapping.
Although the pseudo-labeled core retains only the centroid-nearest 3\% of each
cluster, it covers the major semantic regions of the source dataset. The final
15\% subset similarly remains distributed across the original representation
space instead of collapsing into a few dense modes. This pattern reflects the
complementary effects of proportional cluster-wise allocation and
\begin{figure*}[!h]
    \centering

    \begin{subfigure}[t]{0.48\textwidth}
        \centering
        \includegraphics[width=\linewidth]
        {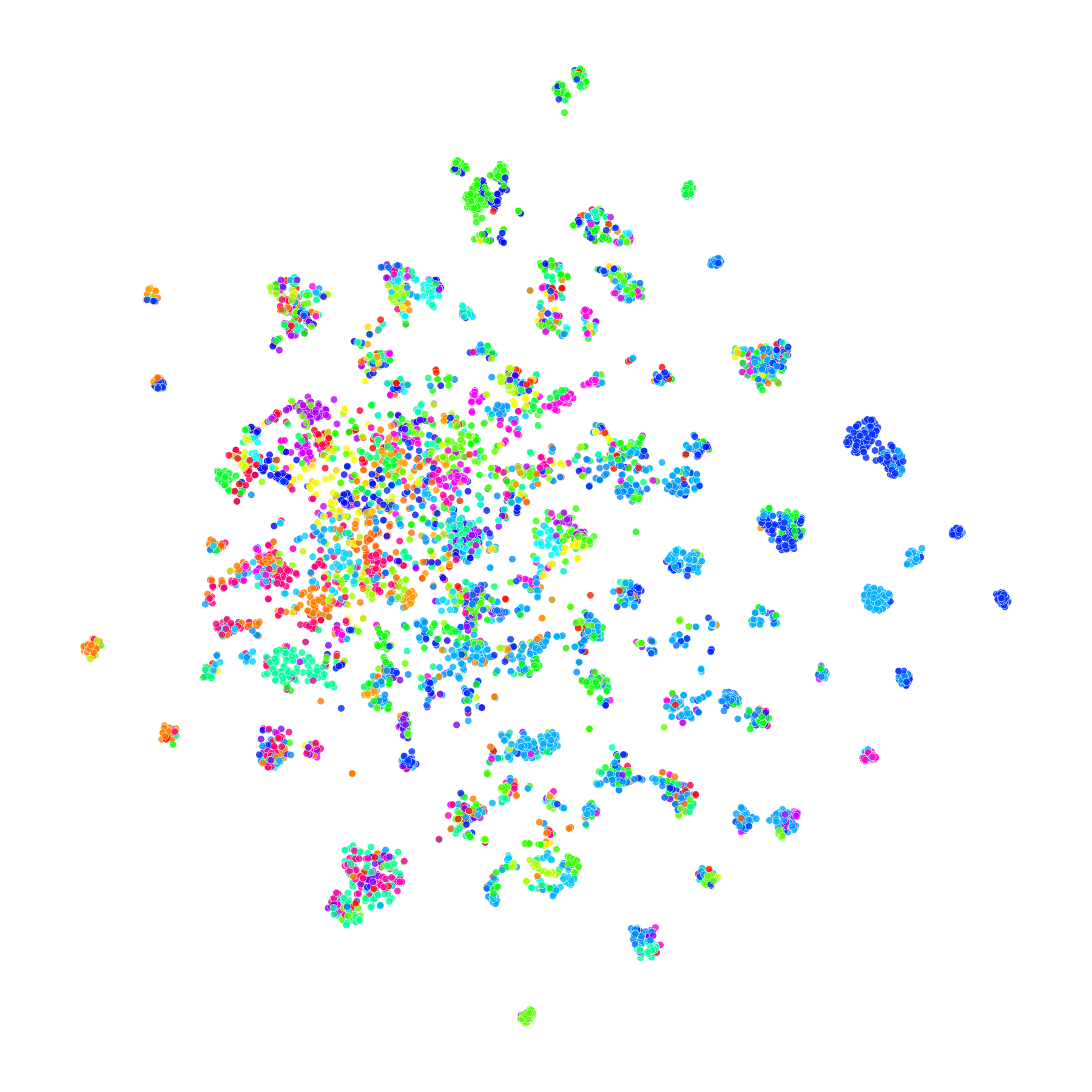}
        \caption{Complete Vision-Flan dataset}
        \label{fig:visionflan_complete}
    \end{subfigure}
    \hfill
    \begin{subfigure}[t]{0.48\textwidth}
        \centering
        \includegraphics[width=\linewidth]
        {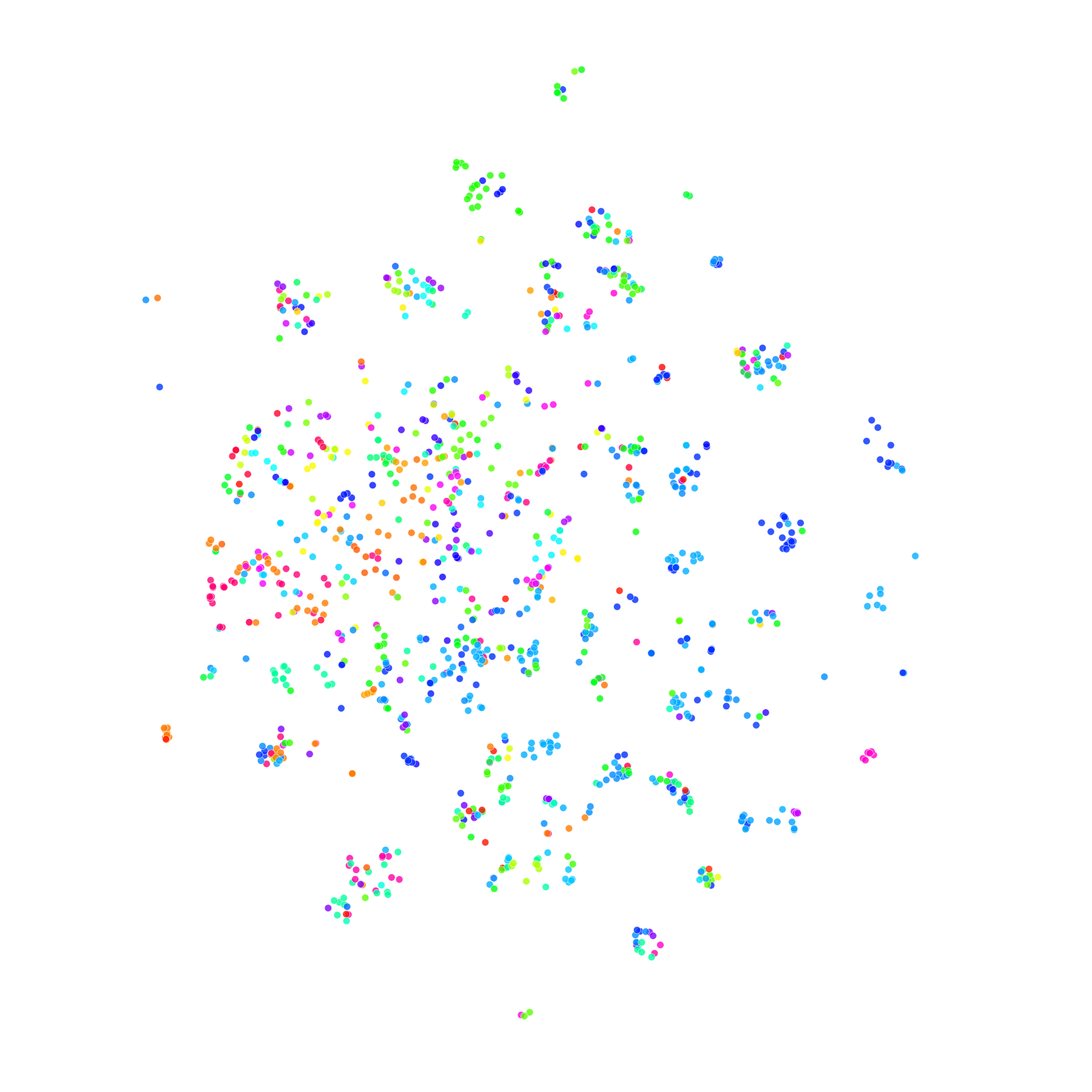}
        \caption{Selected Vision-Flan subset}
        \label{fig:visionflan_selected}
    \end{subfigure}

    \vspace{0.6em}

    \begin{subfigure}[t]{0.48\textwidth}
        \centering
        \includegraphics[width=\linewidth]
        {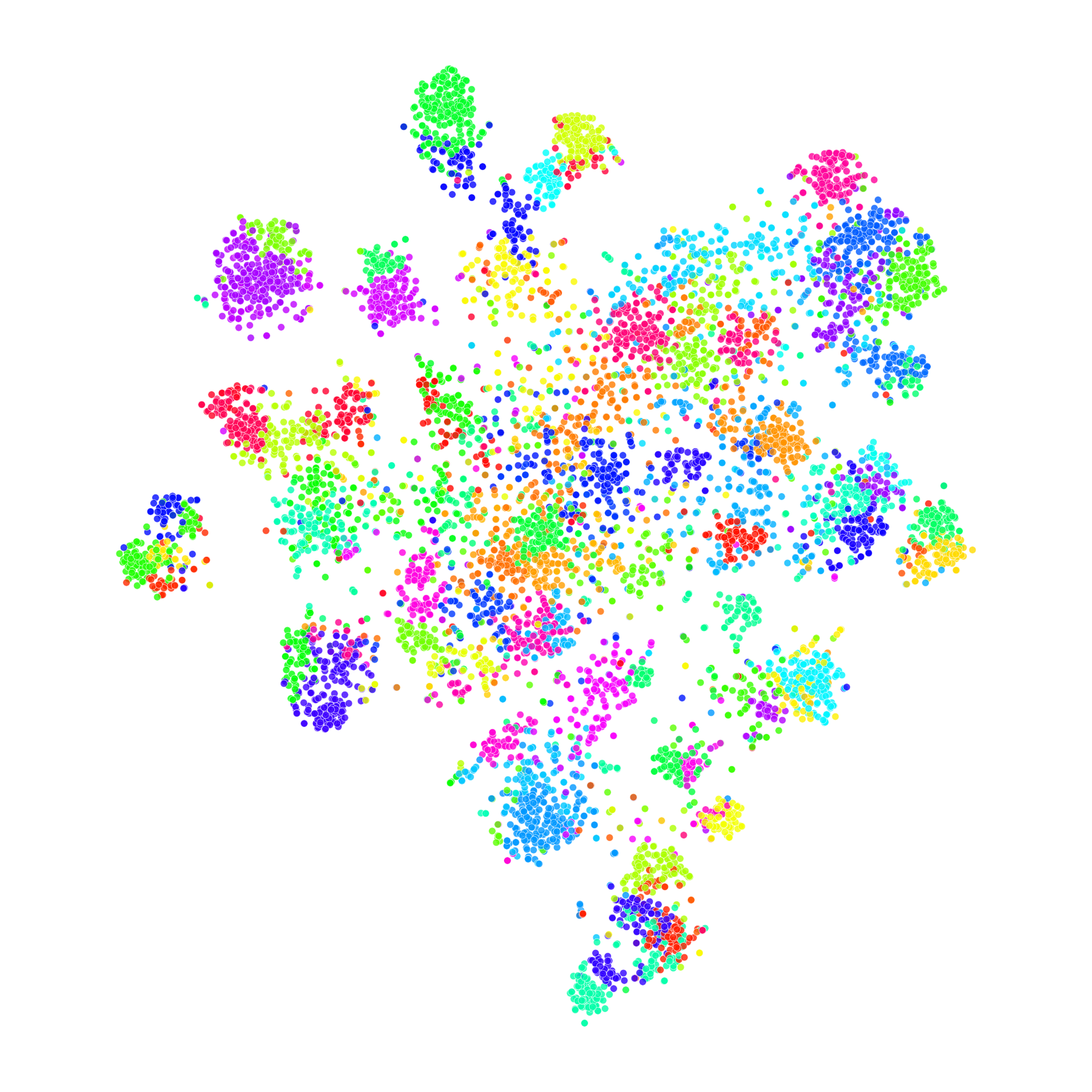}
        \caption{Complete LRV-Sub dataset}
        \label{fig:lrvsub_complete}
    \end{subfigure}
    \hfill
    \begin{subfigure}[t]{0.48\textwidth}
        \centering
        \includegraphics[width=\linewidth]
        {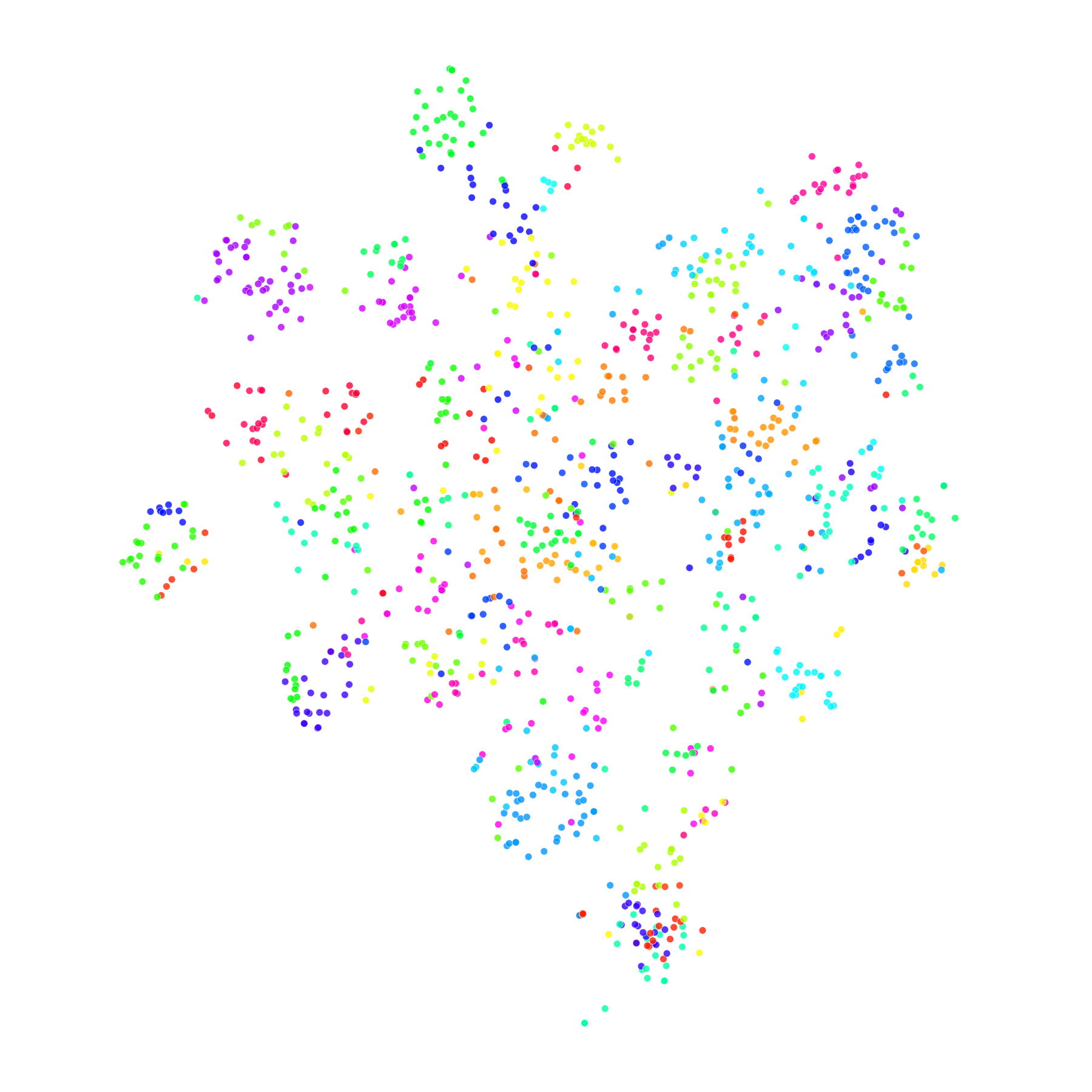}
        \caption{Selected LRV-Sub subset}
        \label{fig:lrvsub_selected}
    \end{subfigure}

    \caption{Distribution of two unseen target datasets before and after
    selection. Each complete--selected pair shares the same t-SNE projection.}
    \label{fig:transfer_visualization}
\end{figure*}
uncertainty-weighted $k$-DPP: the former maintains coarse-grained semantic
coverage, while the latter reduces within-cluster redundancy and promotes
local diversity. The selected subset therefore approximately preserves the
semantic structure of the source distribution while prioritizing potentially
informative samples.

\noindent\textbf{Visualization on Unseen Datasets.}
We further visualize the transfer of the source-trained selector to
Vision-Flan-186K and LRV-Sub-180K. For each target dataset, the complete
dataset and its selected subset share the same t-SNE projection. As shown in
Fig.~\ref{fig:transfer_visualization}, although the selector and semantic
centroids are learned only from LLaVA-625K, the selected subsets continue to
cover the major regions of both target distributions. Dense regions are
substantially compressed, while dispersed regions remain represented. This
provides qualitative evidence that OnceSelect adapts beyond the source-data
distribution: cluster-aware quota allocation maintains coarse semantic
coverage, while within-cluster $k$-DPP reduces local redundancy and promotes
diversity on unseen datasets.

Overall, the visualizations suggest that OnceSelect achieves data reduction
without severe distributional collapse, balancing coarse-grained
representativeness, within-cluster diversity, and sample informativeness.
\end{document}